\documentclass[lettersize, 10 pt, journal]{IEEEtran}
\usepackage[pdftex]{graphicx}
\usepackage{subcaption}

\usepackage{amsmath}
\usepackage{amsfonts}

\IEEEoverridecommandlockouts
\begin{document}
%
\title{\vspace{18pt} \LARGE \bf Shared Control Based on Extended Lipschitz Analysis With Application to Human-Superlimb Collaboration} 

%
%
%
 
\author{Hanjun~Song,~\IEEEmembership{Student Member,~IEEE}, and H.~Harry~Asada,~\IEEEmembership{Life Fellow,~IEEE}%
\thanks{This work was supported by the SUSTech-MIT Mechanical Engineering Education and Research Centers.}%
\thanks{This work involved human subjects or animals in its research. Approval of all ethical and experimental procedures and protocols was granted by the Massachusetts Institute of Technology Committee on the Use of Humans as Experimental Subjects under Application No. 2302000880.}%
\thanks{H. Song (hanjuns@mit.edu) and H. H. Asada (asada@mit.edu) are with the Department of Mechanical Engineering, Massachusetts Institute of Technology, Cambridge, MA 02139, USA. The corresponding author is H. Song.}%
}

%
%

\markboth{IEEE TRANSACTIONS ON ROBOTICS, Regular paper}
{Hanjun Song, H. Harry Asada, \MakeLowercase{\textit{et al.}}: Predictability Analysis in Imitation Learning for Human-Robot Collaboration}
%



\maketitle

\begin{abstract}

This paper presents a quantitative method to construct voluntary manual control and sensor-based reactive control in human-robot collaboration based on Lipschitz conditions. To collaborate with a human, the robot observes the human's motions and predicts a desired action. This predictor is constructed from data of human demonstrations observed through the robot's sensors. Analysis of demonstration data based on Lipschitz quotients evaluates a) whether the desired action is predictable and b) to what extent the action is predictable. If the quotients are low for all the input-output pairs of demonstration data, a predictor can be constructed with a smooth function. 
In dealing with human demonstration data, however, the Lipschitz quotients tend to be very high in some situations due to the discrepancy between the information that humans use and the one robots can obtain. This paper a) presents a method for seeking missing information or a new variable that can lower the Lipschitz quotients by adding the new variable to the input space, and b) constructs a human-robot shared control system based on the Lipschitz analysis. Those predictable situations are assigned to the robot's reactive control, while human voluntary control is assigned to those situations where the Lipschitz quotients are high even after the new variable is added. The latter situations are deemed unpredictable and are rendered to the human.  
This human-robot shared control method is applied to assist hemiplegic patients in a bimanual eating task with a Supernumerary Robotic Limb, which works in concert with an unaffected functional hand.
\end{abstract}

\begin{IEEEkeywords}
Human-Robot Collaboration, Shared Control, Lipschitz Conditions, Supernumerary Robotic Limbs, Predictability, Hidden Markov Model, SuperLimbs 
\end{IEEEkeywords}

%
\IEEEpeerreviewmaketitle

\section{Introduction}
%
%
%
%

\IEEEPARstart{W}{hen} teaching or transferring human skills to robots for human-robot collaboration, a fundamental question arises: which functions should be executed by robots versus humans, and to what extent? This dilemma has a long history of inquiry in the study of human-machine systems \cite{Sheridan1978-et, Endsley1999-he, Parasuraman2000-vr}. Researchers have adopted a taxonomy-based framework to determine the levels of robot autonomy in human-robot interactions \cite{Beer2014-tu}. However, it is still desirable to develop a quantitative methodology that informs autonomy design in human-machine or human-robot systems.

Previous studies about human skill transfer are informative to developing a quantitative method to determine the levels of robot autonomy and human intervention in human-robot collaboration \cite{Asada1989-ek, Asada1991-qg, Liu1992-pv, He1993-pd}. These papers demonstrate that an information discrepancy exists between humans and robots because humans use a variety of information, including subconscious knowledge, especially when performing complex skills, such as dexterous manipulation. The Lipschitz quotients were used to measure the transferability of human skills to robots and to validate the consistency of the data \cite{Asada1991-qg, Liu1992-pv, He1993-pd}. This data analysis method can be applied to determine the levels of robot autonomy and human intervention in human-robot collaboration and shared control. The underlying key concept is that tasks that depend more on information unavailable to robots should be assigned to humans, while those dependent on information available to robots should be assigned to the robots. It is hoped that the quantitative framework can provide fundamental guidelines informing what can be done well by robots and what should be done by humans in collaboration tasks. This could also help reduce conflicts between humans and robots in shared control.

\begin{figure}[t]
\centering
\includegraphics[width=0.5\textwidth]{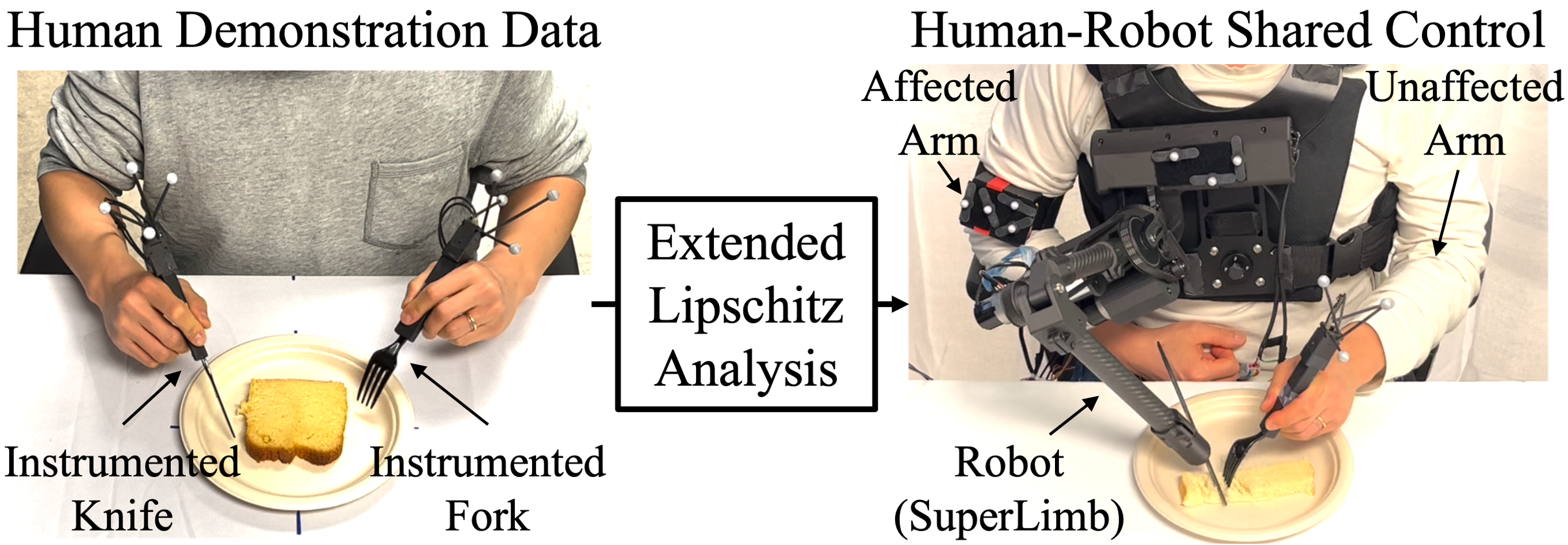}
\caption{Application to hemiplegic patient support in bimanual eating using a knife and fork. The stroke survivor cannot use the right hand for holding a knife. Instead, the SuperLimb holds the knife and coordinates its motion with the fork held by the unaffected hand of the patient.}
\label{first_fig}
\end{figure}

Supernumerary Robotic Limbs, or SuperLimbs for short, are a robotic system where this proposed approach is demonstrated. SuperLimbs are a type of wearable robot that either augments physical capabilities or compensates for lost functionalities. While various designs of SuperLimbs have been developed, such as arms \cite{Parietti2017-lm, Sasaki2017-kq, Vatsal2018-rc, Veronneau2020-ss, Amanhoud2021-rt, Nguyen2019-ky, Bonilla2014-tb, Parietti2014-sa}, fingers \cite{robotics8040102, Cunningham2018-eh, Hussain2017-ew, Prattichizzo2014-tm, Wu2014-qc, Kieliba2021-vb}, and legs \cite{Hao2020-sl, Khazoom2020-jw, Treers2017-ru, Parietti2015-sp, Kurek2017-sm}, they share the problem of determining an appropriate level of robot autonomy and human intervention. So far, two distinct control methods have been explored for SuperLimbs: voluntary and reactive control. Voluntary control corresponds to human intervention and manual operation, while reactive control corresponds to robot autonomy.

Abdi, Burdet, et al. \cite{PLoS-Foot-2015} explored a fully voluntary control method, which exploited the wearer's body movements. Either task-irrelevant movements \cite{Sasaki2017-kq, Kieliba2021-vb, PLoS-Foot-2015, Huang2020-hc, Guggenheim2020-lg} or physiological signals \cite{Parietti2017-lm, Hussain2016-kn} not interfering with the control of natural limbs can be used for voluntary control. As the task becomes more complex, requiring coordination between many Degrees of Freedom (DOF), voluntary control alone is not feasible due to the high cognitive workload. In contrast, Wu and Asada explored a fully reactive control method in which the control commands for the SuperLimb were generated in response to human finger movements based on artificial synergy \cite{Wu2016-xf}. However, the SuperLimb was volatile to unpredictable situations, such as changes in the user's desired behavior, as the users did not have direct control of the SuperLimbs. Therefore, the applications of reactive control are limited to specialized activities, such as overhead panel installation \cite{Bonilla2014-tb}, fuselage assembly \cite{Parietti2014-sa}, grasp support \cite{Setiawan2020-yk}, or body support while performing works near or on a ground \cite{Kurek2017-sm}. 

To overcome these limitations, voluntary control and reactive control must be integrated. The authors have attempted to integrate them by decomposing the motion based on principal component analysis of human demonstration data into a subspace that is predictable and a second subspace that is not predictable \cite{Song2021-dh}. The former is controlled with reactive control, while the latter is controlled by voluntary control. However, there were two limitations to the integration method: 1) the predictability was dependent on a specific predictor model, and 2) the integration was completed at the hardware level, which is task-specific and inflexible. 

The goal of the current work is to establish a quantitative methodology for constructing a human-robot shared control based on predictability analysis that does not depend on a predictor structure. Lipschitz quotients will be used as a quantitative measure for evaluating the existence of a continuous function that predicts a desired output in relation to each input in the data. This analysis will reveal that human demonstration data often include data points, or situations, with high Lipschitz quotients, indicating that there is no single continuous function in the input-output map. The objective of the current work is to extend the Lipschitz-based data analysis in two aspects:
\begin{itemize}
\item  Extend the continuous predictor function to a hybrid discrete-continuous function to improve the predictability. The data is clustered into a set of discrete states, and a continuous predictor is constructed for each discrete state. A Hidden Markov Model (HMM) is used for representing and predicting the discrete nature of state dynamics. While the hybrid discrete-continuous predictor decreases the Lipschitz quotients, in general, confounding situations with high Lipschitz quotients may still exist, which requires intervention, i.e., human voluntary control.
\item Construct a human-robot shared control system based on the extended Lipschitz data analysis above. Predictor-based, reactive control is employed for those situations characterized as low Lipschitz quotients. Those situations of high Lipschitz quotients are rendered to the human: manual, voluntary control. It is expected that the integration of these reactive and voluntary control results in harmonized human-robot shared control with reduced conflict between the two.
\end{itemize}

This method is applied to bimanual eating manipulation using a knife and fork, where the fork is held by the human and the knife is held by the robot as shown in Fig. \ref{first_fig}. This application will be useful for assisting hemiplegic patients' daily eating tasks. Bimanual eating data is collected from three healthy human subjects, and a human-robot shared controller is constructed and implemented on the SuperLimb system. 

\section{Information Discrepancy in Imitation Learning and Regression Predictability}

This section briefly describes the background of Lipschitz-based data analysis and the framing of the problem to be addressed in the current work.

\subsection{Behavioral Cloning}

Behavioral cloning \cite{Pomerleau-1989-15721} is one of the simplest imitation learning methods to teach robots human skills. Behavioral cloning uses a set of expert demonstrations, $\Xi = \{\xi^1, ... ,\xi^D \}$, where each $i$-th demonstration, $\xi^i$, consists of a sequence of observation-action pairs from the beginning to the end of the demonstration:

\begin{equation}\label{eq:demo_set}
    \xi^i = \{(\boldsymbol{o}_0^i, \boldsymbol{u}_0^i), (\boldsymbol{o}_1^i, \boldsymbol{u}_1^i), ... \}, 
\end{equation}
where $\boldsymbol{o}_t^i \in \mathcal{O} \subset \mathbb{R}^d$ and $\boldsymbol{u}_t^i \in \mathcal{U}  \subset \mathbb{R}^l$ denote an observation and action at time $t$ in demonstration $\xi^i$, where the sets of observations and actions are denoted as $\mathcal{O}$ and $\mathcal{U}$, respectively. In this paper, $\Xi$ represents a set of demonstrations from one human subject, and the same methods can be applied to other sets of demonstrations from different human subjects. 

The goal is to learn a policy, $\pi: \mathcal{O} \rightarrow \mathcal{U} $, that defines the following functional relation between the observation and action by imitating the expert: 

\begin{equation}
    \boldsymbol{u}_t = \pi(\boldsymbol{o}_t).
\end{equation}

This can be achieved through supervised learning, where the difference between the predicted action $\pi(\boldsymbol{o}_t^i)$ and the expert action, $\boldsymbol{u}_t^i$, is minimized based on a loss function $L$, as described below:

\begin{equation}\label{eq:sup_learn}
    \hat{\pi} = \arg \underset{\pi}{\min} \sum_{\xi^i \in \Xi,} \sum_{(\boldsymbol{o}_t^i, \boldsymbol{u}_t^i) \in \xi^i} L(\pi(\boldsymbol{o}_t^i),\boldsymbol{u}_t^i),
\end{equation}
where $\hat{\pi}$ is the learned policy. Although there are known theoretical limitations to behavioral cloning, such as the compounding errors caused by a distributional mismatch between learned and expert policies \cite{Ross2011-gj}, there are also compelling results in practice that show robots learning complex skills through behavioral cloning \cite{Zhang2018-dy, Florence2020-mg, florence2021implicit, 9104757}. In the current study, an expert's behaviors are considered desirable actions that the robot should take.

\begin{figure}[t]
\centering
\includegraphics[width=0.43\textwidth]{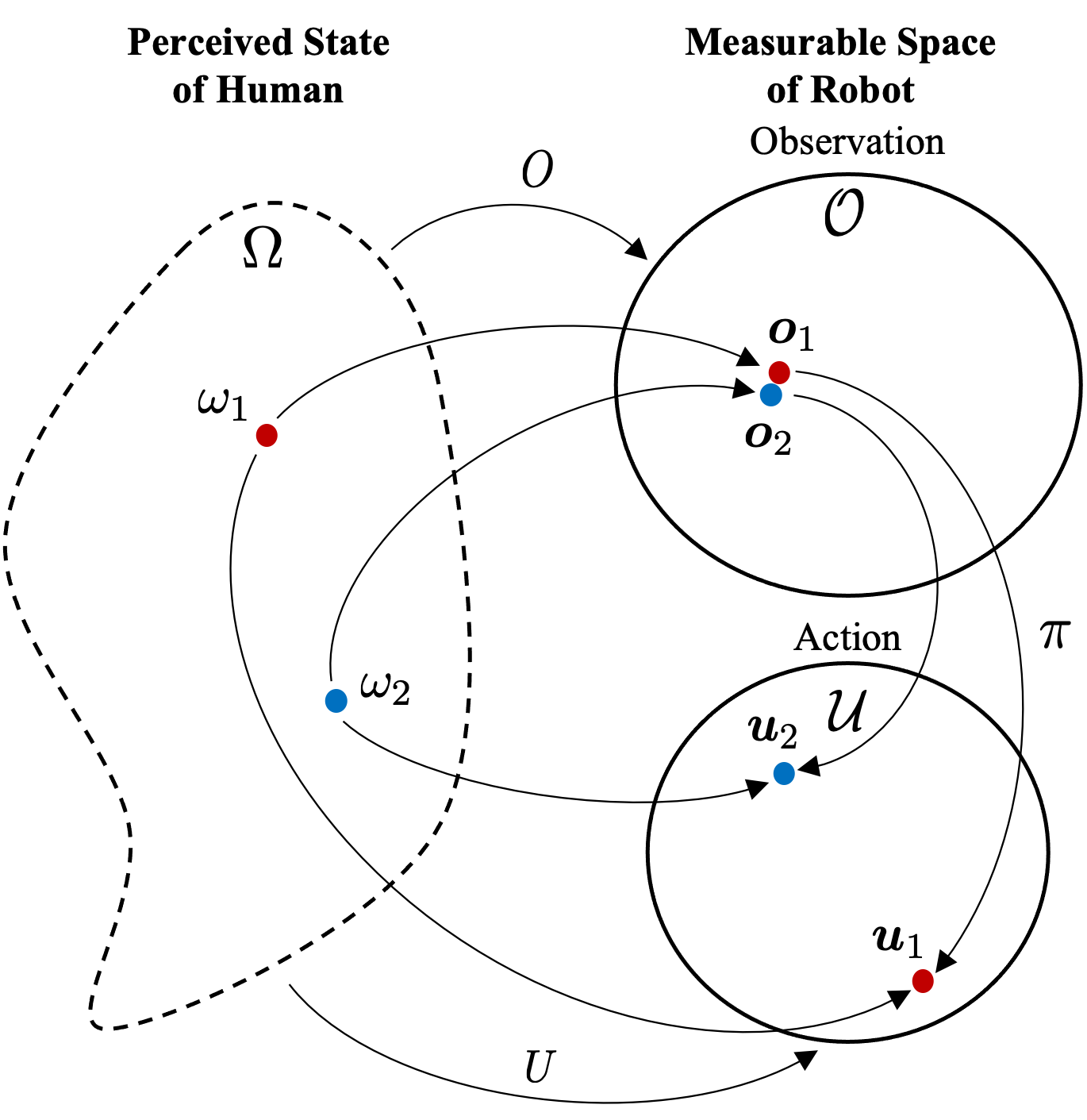}
\caption{The diagram shows how information discrepancy between humans and robots can result in confusion when transferring human skills to robots. Two different states are illustrated as $\omega_1$ and $\omega_2$ in the perceived state of humans, $\Omega$. It may be difficult for robots to learn a correct policy, $\pi$, if the corresponding observations, $\boldsymbol{o_1}$ and $\boldsymbol{o_2}$, are very similar while the corresponding actions, $\boldsymbol{u_1}$ and $\boldsymbol{u_2}$, are very different in the measurable space of robots.}
\label{mapping_diagram}
\end{figure}

\subsection{Fundamental Limitations to Human-Robot Collaboration}
\label{sec:info_disc}
It is important to note that the above imitation learning is fundamentally limited due to the information discrepancy between humans and robots as illustrated in Fig. \ref{mapping_diagram}. When teaching human skills to the robot, only available sensors are used to observe human behaviors to infer the corresponding desired actions. Available sensors, however, are not always able to capture all of the information that humans utilize, that is, the perceived state space of the human $\Omega$. This means that information can only be partially collected through observation in the observable space of robots, $\mathcal{O}$. If a human skill is more dependent on the information that is unavailable to the robot, the robot may not be able to infer the correct human skills during collaboration, and therefore, it would be better to assign those actions to a human's discretion, or voluntary control. We would say predictability is low in this case. On the other hand, if a human skill is more dependent on shared information, the robot would be able to infer the correct human skills based on robot autonomy or reactive control. We would call this a high predictability case. This strategy of identifying information discrepancy will be used to assign the tasks to be controlled either by human intervention or by robot autonomy in human-robot collaborations.

\subsection{Regression Predictability based on Lipschitz Conditions}\label{sec:reg_lip_con}

Lipschitz conditions have been used for assessing predictability in the context of human skill transfer, or imitation learning \cite{Asada1991-qg, Liu1992-pv, He1993-pd}. A function, $f: \mathcal{X} \rightarrow \mathcal{Y}, \mathcal{X} \subset \mathbb{R}^n, \mathcal{Y} \subset \mathbb{R}^m$, is Lipschitz continuous if it satisfies the condition:
\begin{equation} \label{eq:lip_con}
   d_{\mathcal{Y}}(f(\boldsymbol{x_1}),f(\boldsymbol{x_2})) \le Kd_{\mathcal{X}}(\boldsymbol{x_1},\boldsymbol{x_2}) \quad  \forall \boldsymbol{x_1}, \boldsymbol{x_2} \in \mathcal{X},
\end{equation}
for a real constant $K \ge 0$ and metrics $d_{\mathcal{X}}$ and $d_{\mathcal{Y}}$ on the set $\mathcal{X}$ and $\mathcal{Y}$, respectively. The value $K$ is referred to as the Lipschitz constant, and the function can be referred to as the $K$-Lipschitz. The inequality (\ref{eq:lip_con}) is called a Lipschitz condition. Given that a Euclidean distance is used as a metric, the Lipschitz condition (\ref{eq:lip_con}) can be rewritten as follows:

\begin{equation} \label{eq:lip_con_euc}
   \frac{\lVert f(\boldsymbol{x_1})-f(\boldsymbol{x_2})\rVert}{\lVert \boldsymbol{x_1}-\boldsymbol{x_2}\rVert} \le K \quad  \forall \boldsymbol{x_1}, \boldsymbol{x_2} \in \mathcal{X}.
\end{equation}

The quotient on the left-hand side becomes large or unbounded if two points in the input space, $\boldsymbol{x_1}, \boldsymbol{x_2}$, are close to each other, or overlap, and the corresponding outputs, $\boldsymbol{y_1}, \boldsymbol{y_2}$, are significantly different. This is a confounding situation. As illustrated in Fig. \ref{mapping_diagram}, the sensors of the robot are unable to distinguish the two states $\omega_1, \omega_2$, both producing almost the same observations, $\boldsymbol{o_1}, \boldsymbol{o_2}$, while the corresponding expert's actions differ significantly. This makes the quotient large or unbounded. This occurs when the sensor space of the robot is incomplete; some information that the human uses is missing in the robot's sensor space. The human actions are unpredictable in this situation. 

On the other hand, if the Lipschitz quotients are bounded or not excessively large for all the pairs of points in the entire input space, the human actions are predictable with the given sensor measurements. 



\begin{equation} \label{eq:lip_q}
   q(\boldsymbol{o}^*, \boldsymbol{u}^*) = \sup_{\substack{\boldsymbol{o} \in \boldsymbol{O} \setminus \boldsymbol{o}^*, \boldsymbol{u} \in \boldsymbol{U} \setminus \boldsymbol{u}^*}}  \frac{\lVert \boldsymbol{u}-\boldsymbol{u}^*\rVert}{\lVert\boldsymbol{o}-\boldsymbol{o}^*\rVert} \le K,
\end{equation}
where $\boldsymbol{u}$ is the expert's action at observation $\boldsymbol{o}$. Supremum is taken for all the observations except for the point of examination, $\boldsymbol{o} \in \boldsymbol{O} \setminus \boldsymbol{o}^*$, and the corresponding output actions. The point of examination is excluded as denoted by $\setminus$. 
The above point-wise Lipschitz quotient represents the worst case,  the most confusing, confounding case at each point in the observation space. 

Rigorously speaking, this Lipschitz analysis requires the examination of an infinite number of data points continually distributed within the input space, which is infeasible. Thus, the Lipschitz conditions are instead approximately assessed with a finite number of demonstration data. The assumption is that the sampled data is dense enough to represent the expert policy. The above Lipschitz conditions (\ref{eq:lip_q}) are evaluated for the finite number of data in $\Xi$.


In this study, the data points satisfying the Lipschitz conditions are said to be regression-predictable, and the ones not satisfying the condition are said to be regression-unpredictable. An advantage of using Lipschitz quotients is that they are model-independent, which allows us to evaluate data separately from a prediction model.  If the measure of predictability is model-dependent, the quantity reflects the errors originating from both the data and the model.

With this Lipschitz data analysis of human demonstrations, we aim to construct a human-robot shared control system. Situations in the human demonstration where the data points satisfy the Lipschitz conditions are assigned to the robot's reactive control because the robot can predict a desired action properly. On the other hand, those having large Lipschitz quotients are rendered to the human. The human would use information that the robot could not capture, or the human action would be spontaneous, having no identifiable relationship with the robot observation.

The challenge is to fill the gap between the information that the human uses and the one that the robot can observe. This requires finding the missing information or a new variable to augment the robot's sensor space. Confounding cases, like $\boldsymbol{o_1}, \boldsymbol{o_2}$, can be separated by augmenting the robot sensor space. If such additional variables, measurements, or information are found, the Lipschitz quotients will become lower, and a continuous map from $\mathcal{O}$ to $\mathcal{U}$, i.e. policy $\pi$, can be obtained with a standard machine learning technique. If such additional information is not available, or the Lipschitz quotients are still high even after augmenting the robot sensor space, the human expert's actions are deemed not predictable in the regions of confounding cases. As such, the confounding cases can not be handled by the robot; these should be rendered to human voluntary control.

Before addressing these challenging issues, we consider a concrete and practical context of the task and actual data obtained from human subjects.

\begin{figure}[t]
    \centering
    \begin{subfigure}{0.4\textwidth}
        \centering
        \begin{subfigure}[t]{1.0\textwidth}
            \centering
            \includegraphics[width=1.0\linewidth]{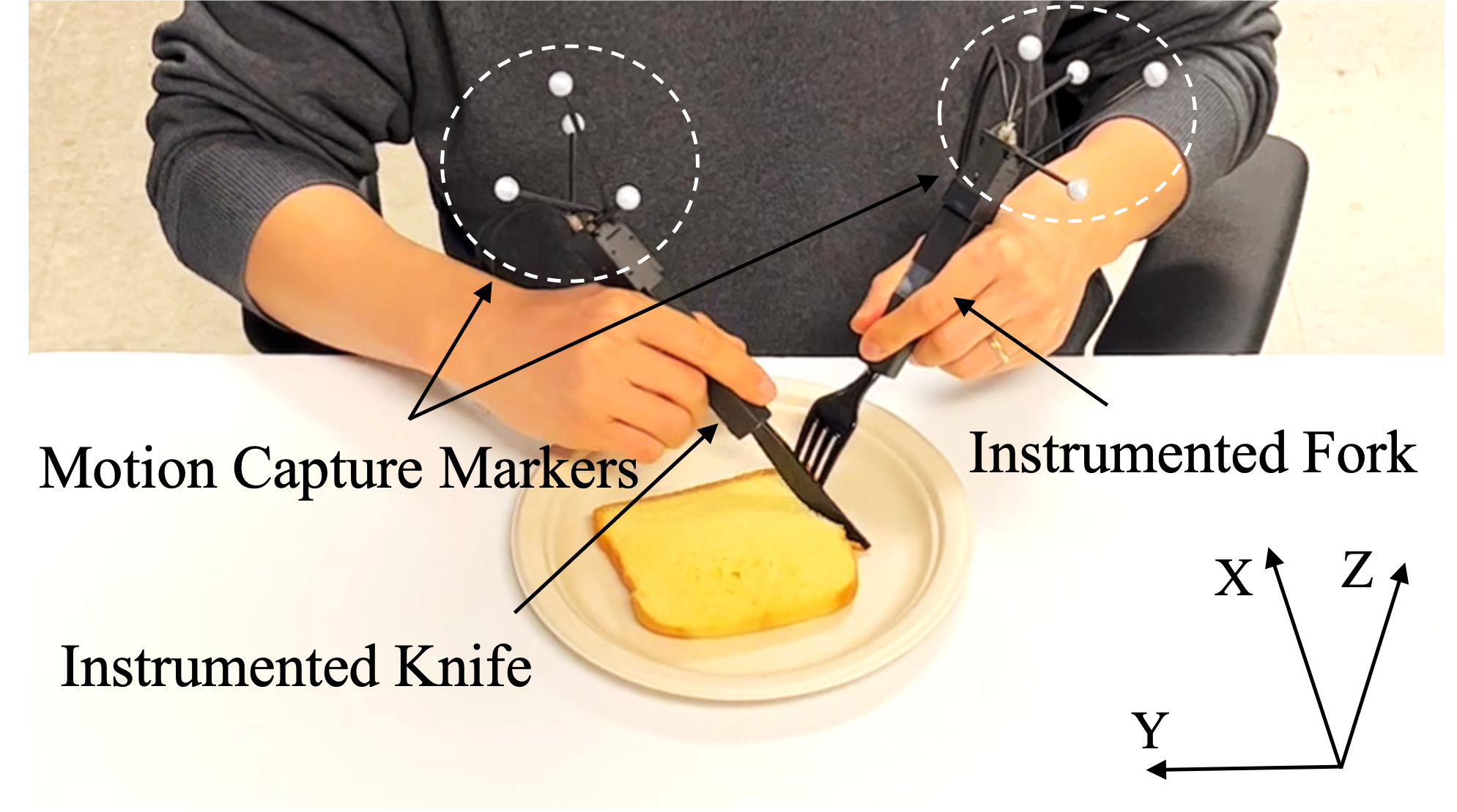}
            \caption{Data collection setup}
            \label{data_col_setup}
        \end{subfigure}
        \begin{subfigure}[t]{1.0\textwidth}
            \centering
            \includegraphics[width=1.0\linewidth]{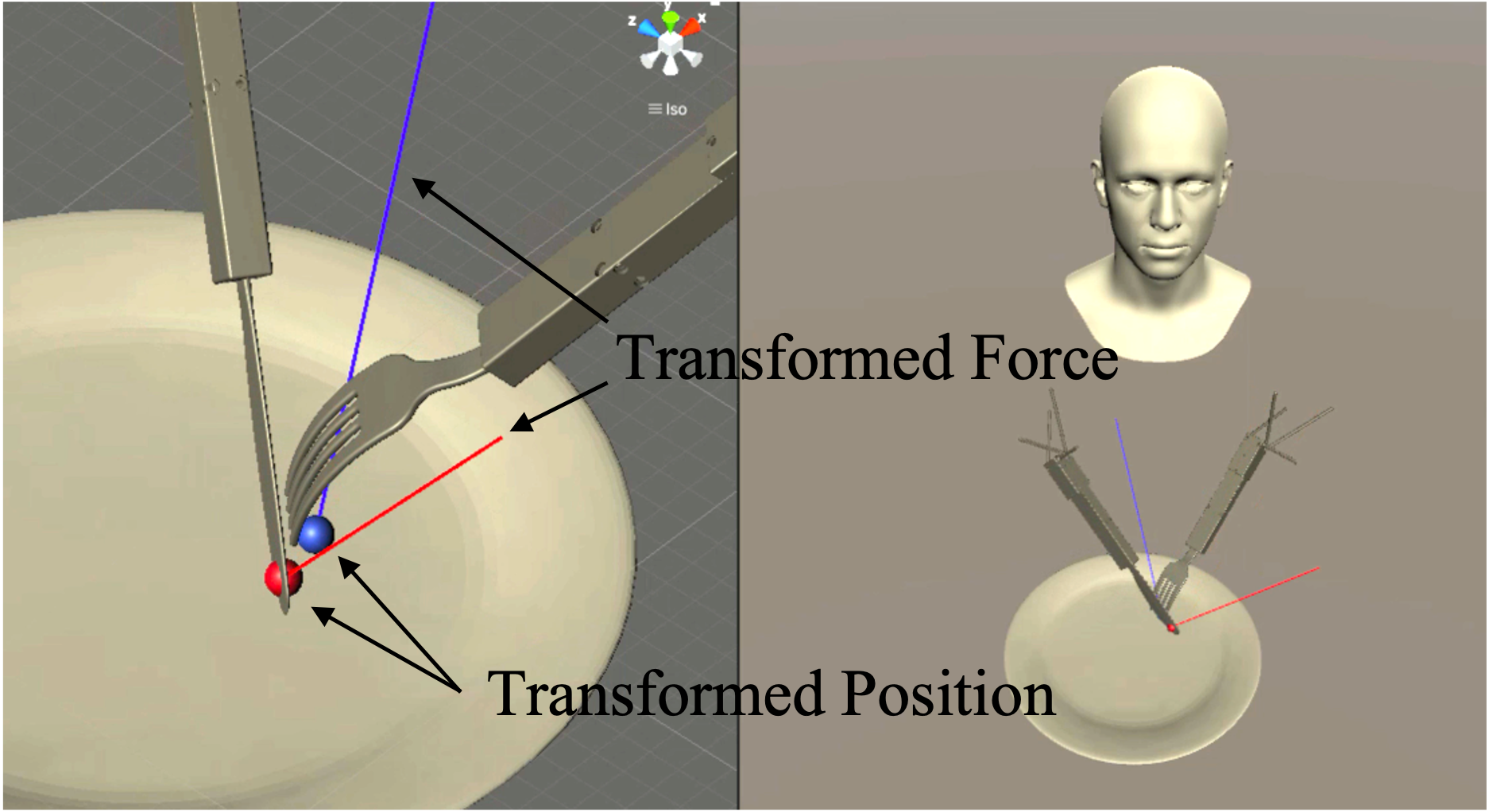}
            \caption{Data validation}
            \label{data_unity}
        \end{subfigure}
    \end{subfigure}
    \caption{Bimanual eating demonstration data is collected using a fork and knife instrumented with motion capture markers and force sensors as shown in (a). Pre-processed data is validated by visualizing it using Unity as shown in (b).}
    \label{data_preproc}
\end{figure}

\begin{figure*}[t]
    \centering
    \includegraphics[width=0.95\linewidth]{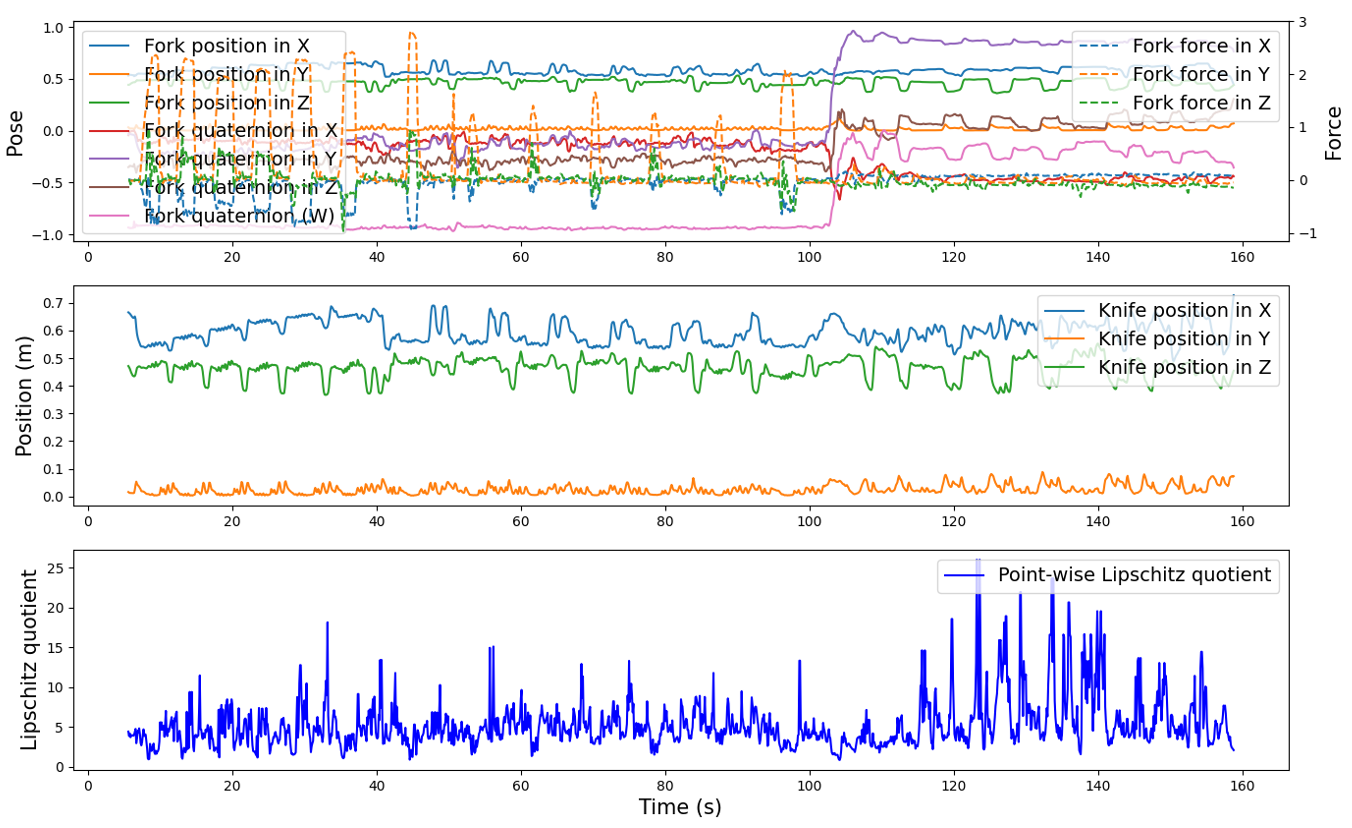}
    \caption{Transformed data with point-wise Lipscthiz quotients}
    \label{data_vis}
\end{figure*}

\section{Task and Data}
The proposed method can be applied to a broad range of problems and settings. However, a specific application has been selected for the current study to demonstrate the method with real data.

\subsection{Bimanual Eating Assistance for Hemiplegic Patients}\label{sec:bi_eat_hemi}
According to American Stroke Association, 90 $\%$ of stroke survivors have some degree of paralysis immediately following the stroke, including hemiplegia, a paralysis that affects one side of the body and results in the inability to control the voluntary movement of a muscle or a group of muscles. Eating difficulties are a particular challenge that hemiplegic patients may experience, and difficulty handling cutlery due to restricted limb movements is a major factor that may contribute to this predicament \cite{Jacobsson1996-pb, Perry2003-rc}. Caregivers can support hemiplegic patients when eating. However, patients have expressed feelings of shame about being dependent on others for feeding and of embarrassment due to a lack of oral control or feeling rushed to finish their meals \cite{Jacobsson2000-so}. Therefore, bimanual eating assistance is a task in which human-robot collaboration can be applied to help hemiplegic patients better control cutlery and thus gain independence during eating. In the current study, it is assumed that a patient has right-sided hemiplegia, and therefore, the human holds a fork with the left hand, which is unaffected, and a robot, the SuperLimb, holds a knife on the right side as illustrated in Fig. \ref{first_fig}. The motion of the knife is then predicted based on the pose and force of the fork under the assumption that the fork takes the lead most of the time in the coordination between the fork and knife. 

\subsection{Data Collection and Preprocessing}\label{sec:data_col_prep}
To find the policy of bimanual eating, $\pi$, using behavioral cloning, sets of expert demonstrations are collected from three healthy human subjects. The data consists of 6 degrees of freedom (DOF) pose and 3 DOF force of a fork and knife. The experiment was conducted based on the protocol approved by the Massachusetts Institute of Technology Committee on the Use of Humans as Experimental Subjects (IRB) Protocol Number 2302000880. The data is collected using the instrumented fork and knife as shown in Fig. \ref{data_col_setup}. Approximately an hour of demonstration is recorded from each of the three human subjects. 

The raw data was pre-processed to convert it to standardized data represented in an appropriate coordinate system. The transformed data is validated by visualizing it, using the visualization software, Unity, as shown in Fig. \ref{data_unity}. Fig. \ref{data_vis} illustrates one of the demonstration data in the transformed coordinates. The transformed data is standardized to make zero-mean and unit-variance data. Finally, the sliding window method is used for augmenting the input space by adding lagged features. In this work, a window size of 1 second was used, which gave the best performance of the policy across the human subjects.

\subsection{Predictability Analysis of Data} \label{sec:lip_analysis}
Predictability analysis of data can be divided into two processes: a) Computing point-wise Lipschitz quotients as shown in (\ref{eq:lip_q}) and b) identifying subsets of data for voluntary and reactive controls. First, point-wise Lipschitz quotients are computed with the pre-processed set of demonstration data. One of the trials in the dataset is plotted with the corresponding Lipschitz quotients in Fig. \ref{data_vis}. We can see that the Lipschitz quotients go up and down, depending on the tasks. Larger and smaller Lipschitz quotients represent lower and higher predictability, respectively. Next, the Lipschitz constant, $K$, needs to be determined in order to examine Lipschitz conditions and divide the data into subsets of voluntary and reactive control.

\begin{figure}[t]
\centering
\includegraphics[width=0.45\textwidth]{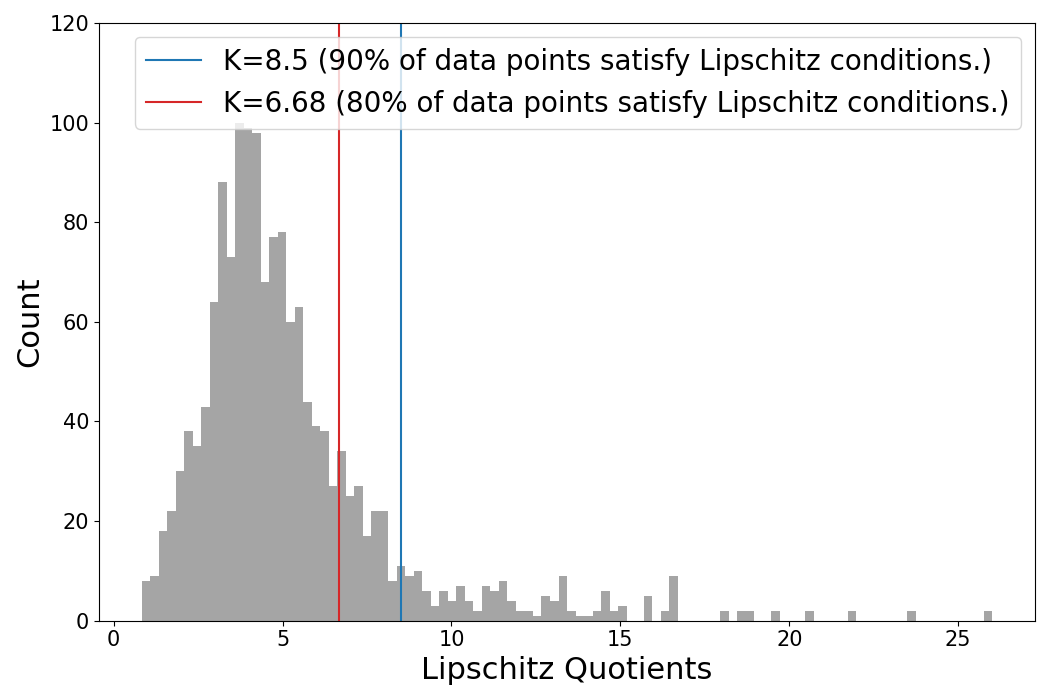}
\caption{Histogram of point-wise Lipschitz quotients and percentile of data points satisfying the Lipschitz conditions.}
\label{lip_thr}
\end{figure}

The Lipschitz constant can be interpreted as a threshold quantity that adjusts the levels of robot autonomy and human intervention. For example, a lower Lipschitz constant assigns more regions of the task to voluntary control and fewer regions to reactive control. Therefore, the levels of human intervention increase, and the levels of robot autonomy decrease. Fig. \ref{lip_thr} shows the histogram of point-wise Lipschitz quotients. If we set the Lipschitz constant $K$ to $8.5$, then $90\%$ of the demonstration data points satisfy the Lipschitz condition. This means that $90\%$ of cases are assigned to reactive control that uses a predictor trained with the $90\%$ of the demonstration data. This also means that $10\%$ of the cases are assigned to the human voluntary control. 
A Lipschitz constant at the $100^{th}$ percentile makes all the data satisfy the Lipschitz condition, and therefore, all the data is used to train the reactive control.

The predictability analysis of data reveals that there are many data points with relatively high Lipschitz quotients. Fig. \ref{data_vis} shows many spikes in the plot of Lipschitz quotients going over, for example, 10. This implies that there are many data points, or situations, that are confounding for a robot. If those situations are all rendered to human voluntary control, the human workload becomes too high, which is not desirable. As such, we must seek additional, useful information that can resolve the confounding situations and improve the predictability of the data in order to ultimately reduce voluntary control efforts.

\section{Task Model and Voluntary-Reactive Control System Design}

In an attempt to improve predictability, we investigate the expert's behaviors that cannot be captured from the observed signals within a fixed time window. Some form of task model or task knowledge could be extracted by examining the entire data set. The human expert may utilize some task knowledge and performs the task in a structured manner based on the model. For example, there exist multiple subtasks in bimanual eating, such as cutting, moving, collecting, etc, and humans may vary the control policy depending on the subtasks. In the previous section, predictability was measured based on the point-wise Lipschitz quotients, which used observations within the specific length of time window. However, task knowledge is a different type of information that we can extract beyond the data in the time window. Extracting an implicit task model and estimating the state of the mode could improve the predictability of data. 

\subsection{Mode classification with Hidden Markov Model}
We hypothesize that there exist multiple discrete modes and probabilistic transitions between them in bimanual eating as a nature of the task. Human subjects were likely to transition to repositioning food items on the dish after cutting them. 
Following cutting actions, the human subjects were likely to collect pieces of food towards the fork for scooping them. This kind of task mode transition can be modeled as a discrete-state, stochastic process, and HMM is used as a model in this study. In HMM, the implicit mode information is treated as hidden states and the transitions between hidden states are assumed to have the form of a (first-order) Markov chain. An HMM is completely defined by $A$, $B$, and $\rho$ denoted by $\lambda = (A,B,\rho)$, where $A$ is a transition probability matrix, $B$ is an observation probability density vector, and $\rho$ is an initial state distribution \cite{Rabiner1989-me}. When there are N number of distinct hidden states, $s_1, s_2, ... , s_N$, the matrix $A = \{a_{ji}\}$ is N by N with

\begin{equation} \label{A_mat}
    \begin{aligned}
        a_{ji} = P(X_t = s_j | X_{t-1} = s_i),
    \end{aligned}
\end{equation}
where $X_t$ is the state at time t and $A$ is row stochastic, meaning that the sum of elements of each row is 1, that is, each row is a probability distribution. 

In our study, observations are continuous signals. As such, the observation process is described with a vector function $B =\{b_j(\boldsymbol{o})\} \in \mathbb{R}^N$ consisting of probability density functions (pdf) conditioned by discrete state $X_t$
\begin{equation} \label{pi_vec}
    \begin{aligned}
        b_j(\boldsymbol{o}) = p(O_t=\boldsymbol{o}|X_t=s_j) ,
    \end{aligned}
\end{equation}
We assume Gaussian distribution for the conditioned distribution.
\begin{equation}
    \boldsymbol{o} | X \sim \mathcal{N}(\boldsymbol{o} | \boldsymbol{\mu_{j}},\boldsymbol{\Sigma_{j}})
\end{equation}
where $\boldsymbol{o}$ is an observation in $\mathcal{O} \subset \mathbb{R}^d$, and $\boldsymbol{\mu_{j}} \in \mathbb{R}^d$ and $\boldsymbol{\Sigma_{j}} \in \mathbb{R}^{d \times d}$ are a mean vector and covariance matrix at state $s_j$, respectively. Lastly, the initial state distribution $\rho = \{\rho_i\}$ is an N-dimensional vector with

\begin{equation} \label{pi_vec}
    \begin{aligned}
        \rho_i = P(X_0=s_i).
    \end{aligned}
\end{equation}

In this study, there are two problems to be solved using HMM: 1) Given an observation sequence, find the model $\lambda = (A,B,\rho)$, and 2) Given $\lambda = (A,B,\rho)$ and an observation sequence, estimate the optimal sequence of hidden states. The first problem can be solved based on a maximum likelihood estimation (MLE) of the HMM model \cite{Bilmes1998-bq}. A standard algorithm for training HMM is forward-backward, or Baum-Welch algorithm \cite{Baum1972AnIA}, which is a special case of the Expectation-Maximization (EM) algorithm \cite{Dempster1977-ze}. The second problem can be solved with a maximum a posteriori (MAP) estimation or Viterbi algorithm. MAP estimation finds the most likely sequence of hidden states given the observation. Viterbi algorithm is an efficient method for MAP estimation of the hidden states based on dynamic programming \cite{Viterbi1967-vu} and is used for this work.

\begin{figure*}[t]
    \centering
    \includegraphics[width=0.95\linewidth]{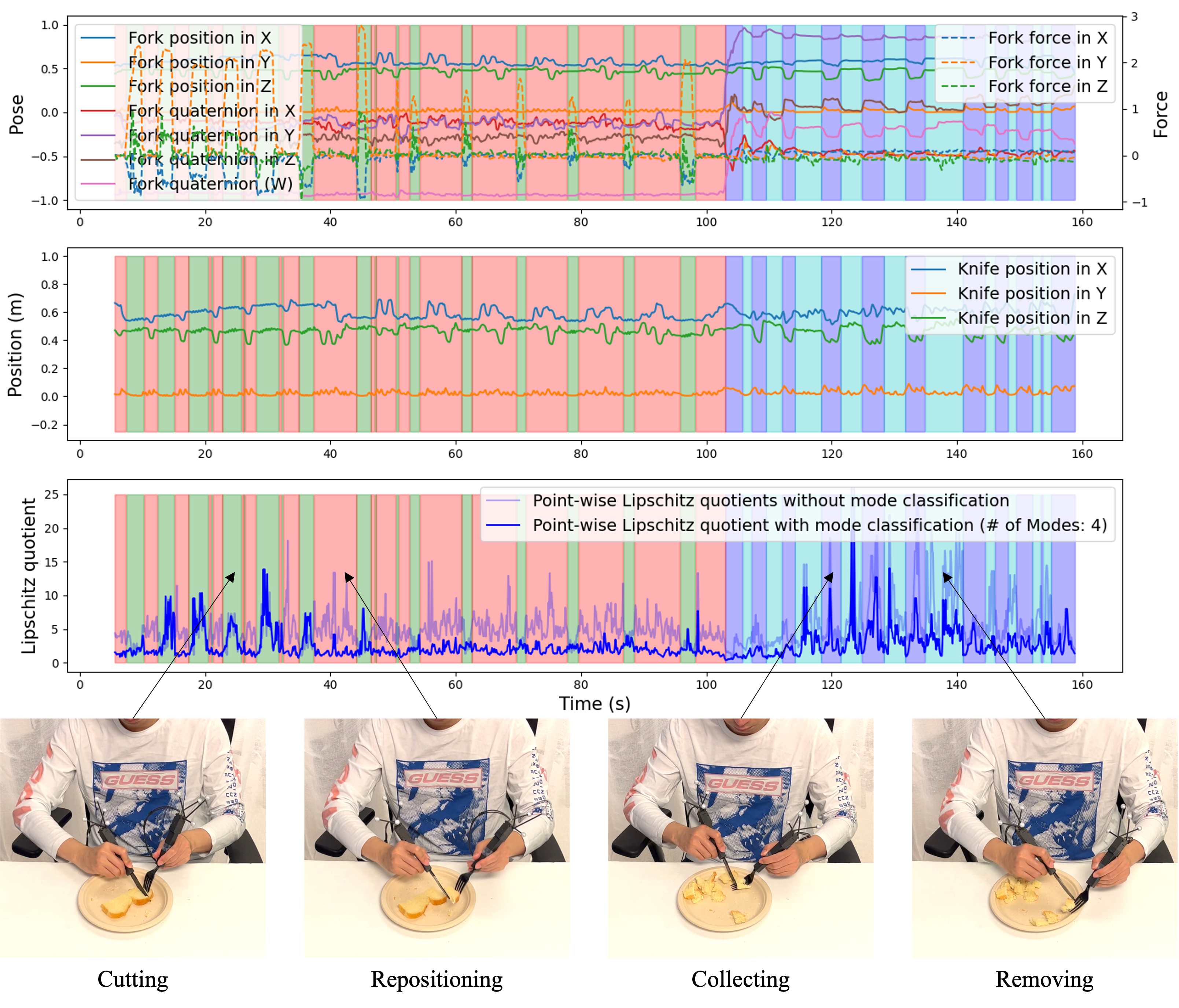}
    \caption{Mode classification using HMM and the corresponding actions.}
    \label{hmm_mode_scene}
\end{figure*}

\subsection{Regression Predictability with Task Knowledge} \label{sec:pred_w_task}
Once the number of modes is given, the HMM model $\lambda = (A,B,\rho)$ can be learned from the sequence of observations in an unsupervised manner. This means that no ground truths of the modes are provided with the observations when learning the HMM model. Fig. \ref{hmm_mode_scene} shows the result of mode classification when the number of modes is four. Interestingly, the four modes correspond to cutting, repositioning, collecting, and removing, even though the ground truths for the modes were not given during the training. The agreement between the manually-labeled modes and the ones based on the HMM model is around 95$\%$. This shows that HMM successfully learned the task model through the data. 

\begin{figure}[t]
\centering
\includegraphics[width=0.35\textwidth]{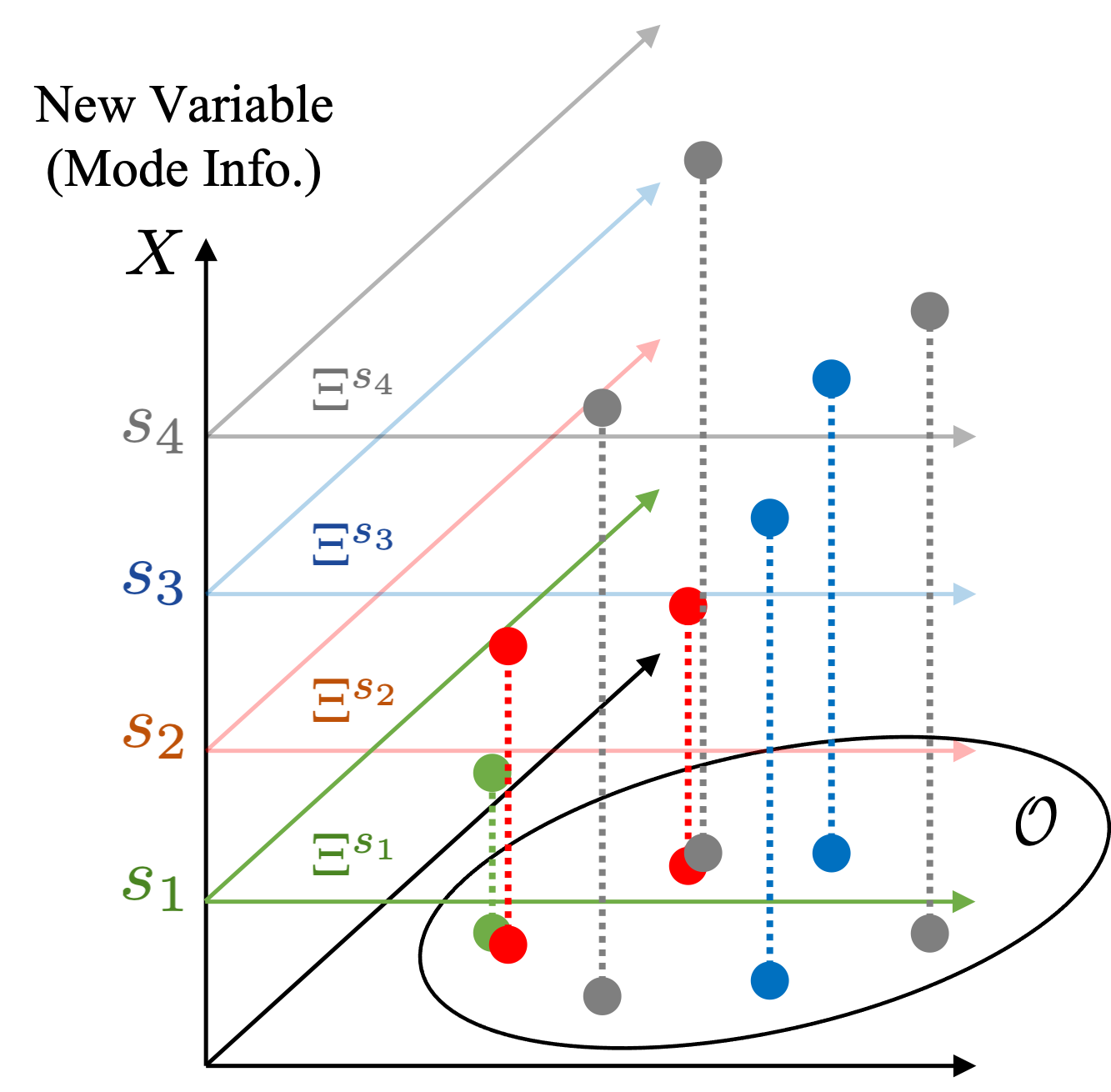}
\caption{Confounding data points can be separated by adding mode information.}
\label{fig:mode_decomp}
\end{figure}

The next step is to recompute the Lipschitz quotients within each mode and investigate if the predictability has improved. First, the dataset is divided into N subsets of the data when the given number of states is N. As shown in Fig. \ref{fig:mode_decomp}, the input data points lie in the original observation space $\boldsymbol{O}$. Now a new axis, the discrete state $X$, is added to the input space, and the individual data points are sorted out and projected onto the four planes, as shown in the figure. Note that some confounding input data points are separated along the new axis $X$.

Let $\Xi^{s_j}$ be the subset of the data projected to the discrete state $s_j$, as shown in Fig. \ref{fig:mode_decomp}. 

\begin{equation} \label{eq:data_mode}
    \begin{aligned}
        &\Xi^{s_j} = \{(\boldsymbol{o}_t^i, \boldsymbol{u}_t^i) \in \Xi \mid X_t^i = s_j \},
    \end{aligned}
\end{equation}
where $X_t^i$ is the hidden state of the data pair $(\boldsymbol{o}_t^i, \boldsymbol{u}_t^i)$. Then, point-wise Lipschitz quotients within each state, $s_j$, can be computed based on (\ref{eq:lip_q}) as follow:

\begin{equation} \label{eq:lip_q_mode}
   q_{s_j}(\boldsymbol{o}^*, \boldsymbol{u}^*) = \max_{\substack{(\boldsymbol{o}, \boldsymbol{u}) \in \Xi^{s_j} \setminus (\boldsymbol{o}^*, \boldsymbol{u}^*)}} \frac{\lVert \boldsymbol{u}-\boldsymbol{u}^*\rVert}{\lVert\boldsymbol{o}-\boldsymbol{o}^*\rVert}.
\end{equation}

Fig. \ref{hmm_mode_scene} shows that the Lipschitz quotients (dark blue) have significantly reduced in most of the regions compared to the original one (light blue). However, this may not necessarily indicate a meaningful improvement in predictability because any division of the dataset, including a random segmentation, can reduce the Lipschitz quotients. To verify that the segmentation of the dataset based on the HMM model is meaningful, the Lipschitz quotients are computed for both randomly segmented data and the one based on the HMM, both having the same number of divisions. Fig. \ref{hmm_dist} assures that there is a significant improvement in predictability for the HMM-based segmentation because the distribution of Lipschitz quotients (light blue) substantially shifted to the left, compared to the one with random segmentation (pink). This can be statistically validated through a t-test. 

Now, it is important to pay attention to those data points where the Lipschitz quotients have not decreased despite the added mode information. If the predictability does not improve even after the efforts of adding more information to the regressors, it is reasonable to ask humans to execute those unpredictable actions rather than generate erratic actions based on the unreliable predictor. 

\subsection{Identification of voluntary and reactive control} \label{sec:vol-re}

Based on the regression predictability enhanced with task mode estimation, the demonstration data of each mode, $\Xi^{s_j}$ described in (\ref{eq:data_mode}), can be divided into two subsets of data based on the Lipschitz condition as follows:

\begin{equation} \label{eq:data_mode_vol_re}
    \begin{aligned}
        &R^{s_j} = \{(\boldsymbol{o}_t^i, \boldsymbol{u}_t^i)\mid q_{s_j,t}^i \le K, \: \text{for} \: (\boldsymbol{o}_t^i, \boldsymbol{u}_t^i) \in \Xi^{s_j}\},\\
        &V^{s_j} = \{(\boldsymbol{o}_t^i, \boldsymbol{u}_t^i)\mid q_{s_j,t}^i > K, \: \text{for} \: (\boldsymbol{o}_t^i, \boldsymbol{u}_t^i) \in \Xi^{s_j}\},
    \end{aligned}
\end{equation}
where $q_{s_j,t}^i$ is the point-wise Lipschitz quotient for the observation-action pair, $(\boldsymbol{o}_t^i, \boldsymbol{u}_t^i)$, within mode $s_j$ computed based on (\ref{eq:lip_q_mode}). $R^{s_j}$ and $V^{s_j}$ represent the subsets of observation data assigned to reactive and voluntary control, respectively. The Lipschitz constant, $K$, can be determined based on the percentile of the distributions of Lipschitz quotients as described in \ref{sec:lip_analysis}.

The reasoning behind this division of data is that unpredictable human skills, not satisfying the Lipschitz conditions, are better controlled by the human or voluntary control because the data is confounding and it is unlikely that the predictor can make a reliable estimate. On the other hand, predictable human skills, satisfying the Lipschitz conditions, can be controlled by robot autonomy or reactive control because the robot has enough information to predict the correct actions. 

It is expected that the division of the control based on predictability could effectively handle and relieve potential conflicts when blending human and robot control inputs in human-robot collaboration. Conflicts between human intervention and robot autonomy in shared control of human-robot collaboration are critical issues. This paper provides a quantitative framework based on the Lipschitz quotients that could manage conflicts effectively based on regression predictability. As discussed in the previous section, the basic idea is to give control to humans in difficult situations for the robots, where Lipschitz quotients are high. On the contrary, a task with low Lipschitz quotients indicates higher predictability, and therefore, the task can be autonomously controlled by the robots. 

\subsection{Integration of voluntary and reactive control}

The basic idea of integrating voluntary and reactive control is to weigh each control based on predictability when summing them up.
However, predictability can be computed only for offline data. For real-time control, predictability must be estimated in real-time. To this end, we construct a binary classifier that can determine whether the current state producing observation $\boldsymbol{o}_t^i$ is predictable or not. First, we define the following binary labels :

\begin{figure}[t]
\centering
\includegraphics[width=0.48\textwidth]{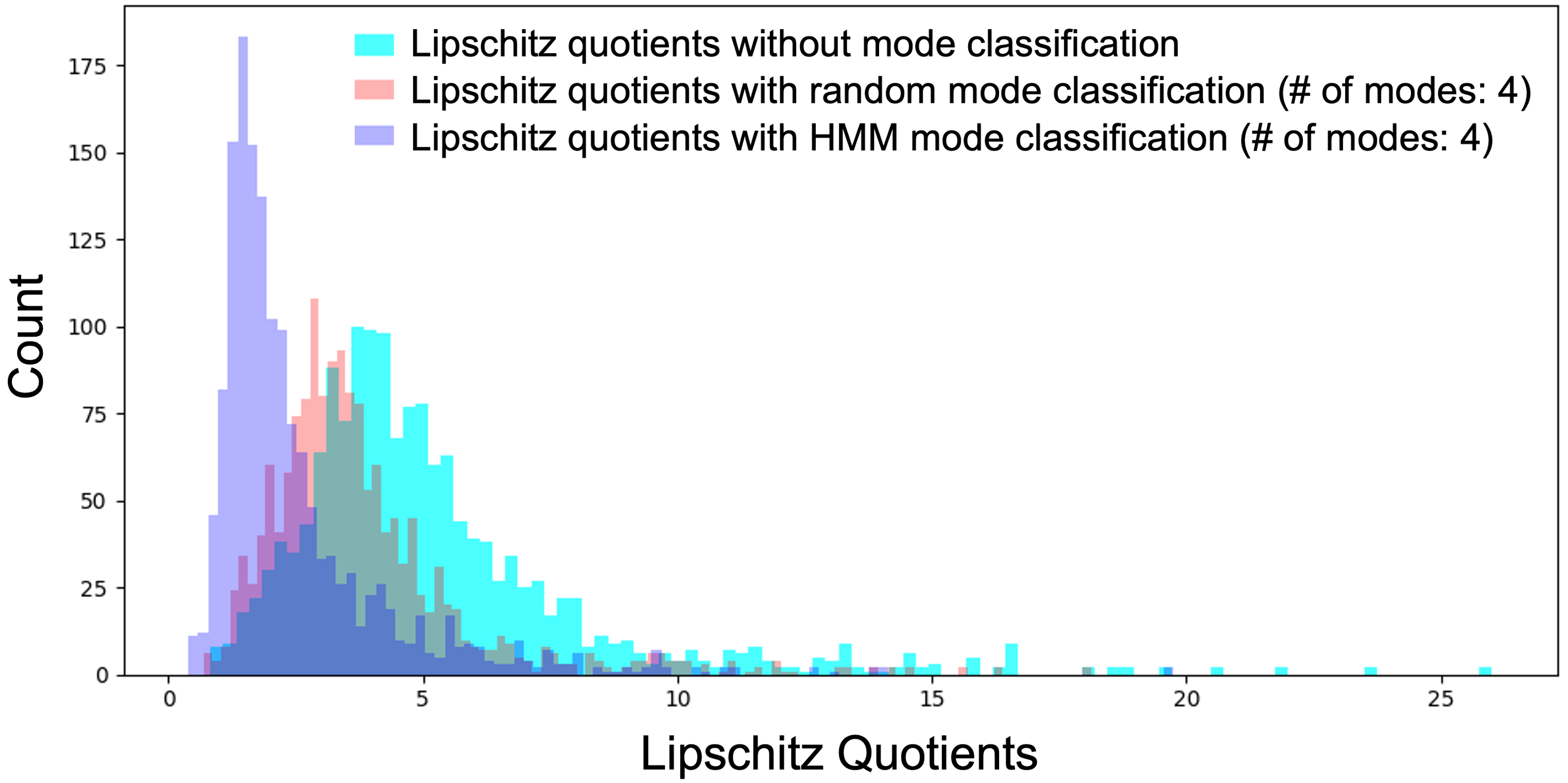}
\caption{Distributions of Lipschitz quotients show that HMM model learned the task knowledge and significantly reduced the Lipschitz quotients.}
\label{hmm_dist}
\end{figure}

\begin{equation}\label{eq:clf}
    Z_{s_j,t}^i= 
    \begin{cases}
        1,& \text{if } q_{s_j,t}^i \le K, \forall (\boldsymbol{o}_t^i, \boldsymbol{u}_t^i) \in \Xi^{s_j}\\
        0,& \text{if } q_{s_j,t}^i > K, \forall (\boldsymbol{o}_t^i, \boldsymbol{u}_t^i) \in \Xi^{s_j}
    \end{cases}.
\end{equation}

Next, a binary classifier $h_j: \mathcal{O} \rightarrow \{0, 1\}$ can be trained for each mode, $s_j$, based on the labels created using (\ref{eq:clf}). Standard machine learning techniques, such as Support Vector Machine, can be used for generating the binary classifier. Finally, voluntary and reactive control can be integrated as a weighted sum of reactive control, $\hat{\pi}^*(\boldsymbol{o})$, of each mode $s_j$ and voluntary control, $\boldsymbol{u}_{v}$, as follows:

\begin{figure*}[t]
\centering
\includegraphics[width=1.0\textwidth]{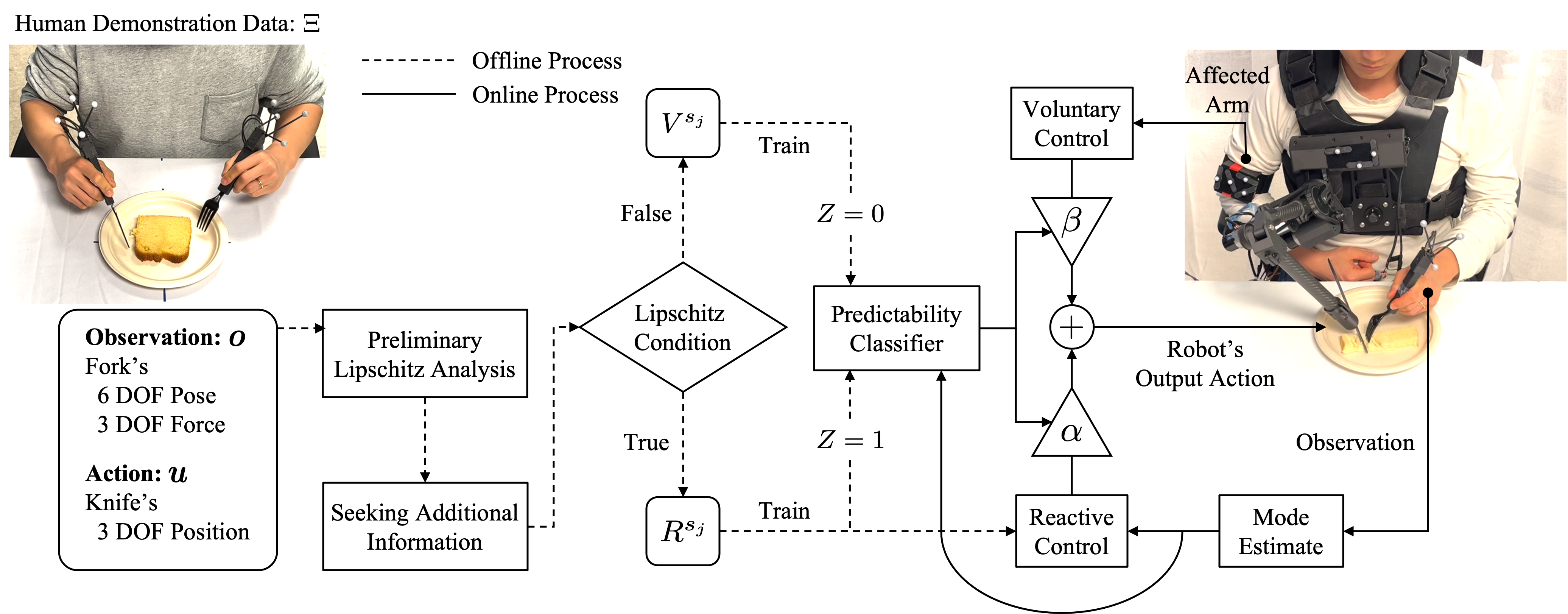}
\caption{Workflow of integrating voluntary and reactive control based on extended Lipschitz analysis. This method requires both offline (dashed lines) and online (solid lines) processes. Human demonstration data, $\Xi$, is divided into two subsets of the data for each mode $s_j$, $R^{s_j}$ and $V^{s_j}$, based on the Lipschitz conditions. Then, the robot autonomy, $\pi$, is trained using $R^{s_j}$. The weighted sum of the reactive control and the voluntary control based on predictability is used as the control input of the robot.}
\label{int_vol_re}
\end{figure*}

\begin{equation}\label{eq:int_vol_re}
    \hat{\boldsymbol{u}} = \alpha\hat{\pi}^*(\boldsymbol{o}) + \beta\boldsymbol{u}_{v},
\end{equation}
where $\alpha$ and $\beta$ are weights of reactive and voluntary control, respectively, determined based on the estimated predictability as follows:

\begin{equation}
    \alpha= 
    \begin{cases}
        1,& \text{if } h_j(\boldsymbol{o})=1\\
        0,& \text{otherwise}
    \end{cases},
\end{equation}

\begin{equation}
    \beta= 
    \begin{cases}
        [0,1],& \text{if } h_j(\boldsymbol{o})=1\\
        1,& \text{otherwise}
    \end{cases}.
\end{equation}

Note that, when predictable, $h_j(\boldsymbol{o})=1$, this integrated policy still allows the human to make an adjustment to the robot control. This is to compensate for the errors of the reactive control and the value is set between 0 and 1 depending on the user's preference. Control command $\hat{\boldsymbol{u}}$ in (\ref{eq:int_vol_re}) is the final integrated action to be fed into the robot's controller. The overall workflow of identification and integration of voluntary and reactive control is summarized in Fig. \ref{int_vol_re}.

\section{Prototyping and Implementation}
To implement the shared control method, a SuperLimb system for assisting a hemiplegic patient is constructed.
\subsection{Hardware Design}
\begin{figure}[t]
\centering
\includegraphics[width=0.4\textwidth]{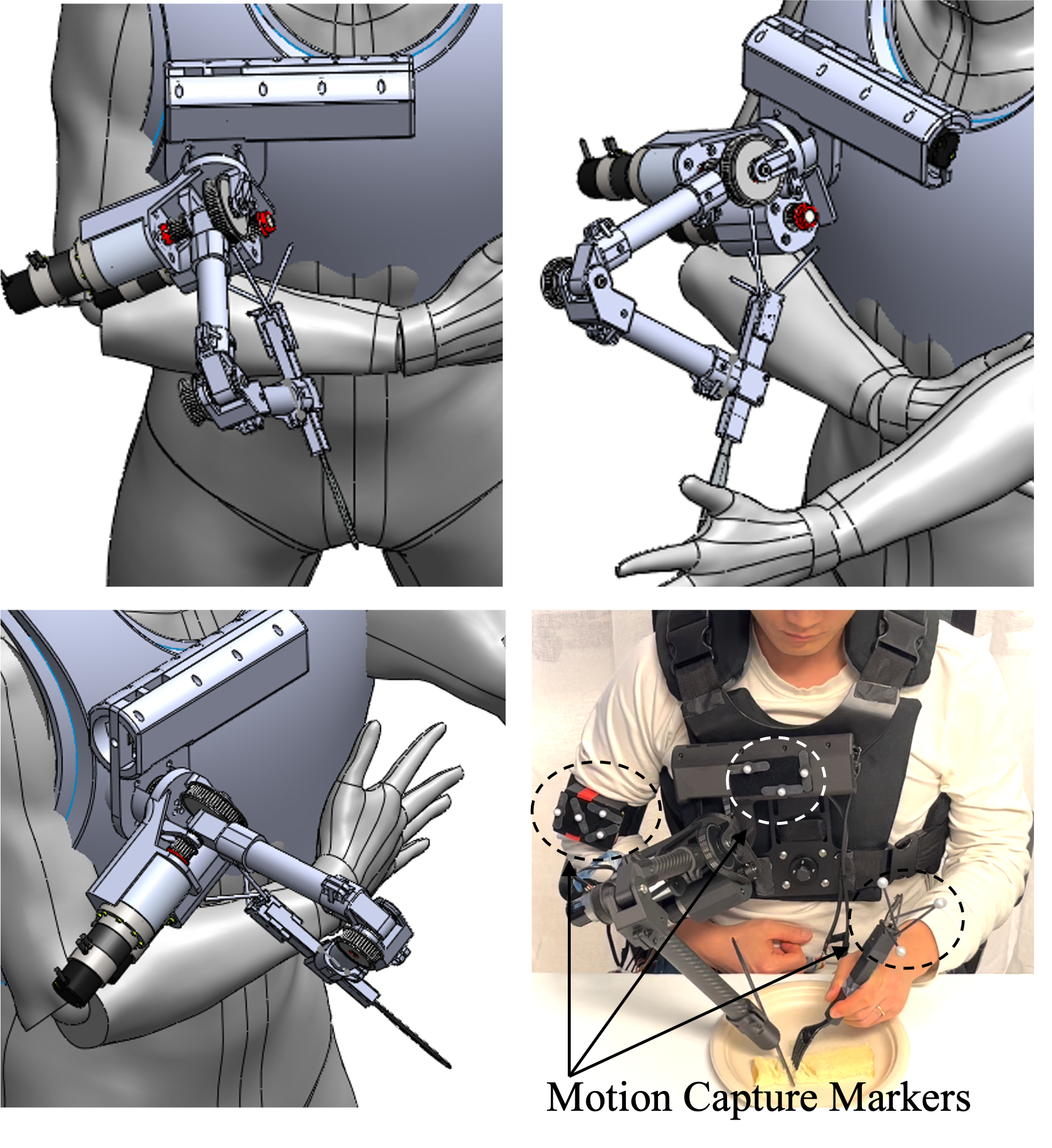}
\caption{Hardware design for the SuperLimb}
\label{fig:hardware}
\end{figure}

A 3 DOF SuperLimb for manipulating a knife is designed and built as shown in Fig. \ref{fig:hardware}. The 3 DOF manipulator is attached to the vest with a solid base so that the human can wear it on the torso. Harmonic Drive actuators (RSF-14B-100) are used with belt and pulley mechanisms. The base joint used a right-angle drive mechanism to place the center of mass of the base as close as possible to the torso. Actuators of the shoulder and elbow joints are placed near the base with the pulley gear ratio of 1:3 and 1:2, respectively. A Raspberry Pi 4 is used with ODrive motor controllers to process incoming data and control actuators. A motion capture system is constructed using OptiTrack Flex 3 cameras. Motion capture markers are attached to the base of the SuperLimb to track and compensate for the movement of the torso. Another set of motion capture markers is attached to the upper arm of the affected arm to generate voluntary control commands. 

Utilizing the elbow motion is not only for generating voluntary commands but also for rehabilitation purposes. Stroke survivors are encouraged to move their affected arm as much as they can. Although the wrist and fingers are slow to recover, many patients can regain the shoulder joints to some extent within the first few months of rehabilitation training \cite{Reeves1975-st, Lindgren2007-oq}. We exploit those shoulder movements for getting voluntary control intention of the wearer. 

\subsection{Software Architecture}
The software of the robot runs off of Robot Operating System (ROS) on Ubuntu 20.04. More specifically, hybrid ROS1-ROS2 is used based on the bridge communication between the two because the motion capture system and ODrive controller run on ROS2 and the force sensor node runs on ROS1. The sampling interval is 10 ms (100 Hz) for the data acquisition and the robot position control. The data was down-sampled to 100 ms (10 Hz) for data processing.

\subsection{Data Processing and Training Robot Autonomy}
As briefly described in section \ref{sec:data_col_prep}, both knife and fork are instrumented with a motion capture system and 3-axis force sensors. Data are pre-processed before evaluating Lipschitz quotients and predictability. First, the down-sampled data are transformed from the coordinate system of the motion capture system and the force sensors to a task coordinate system at the tip of the knife and fork. The data are standardized with unit magnitudes of the individual sensors. A time window is then introduced to expand the input space by concatenating the sensor readings over the time window. An optimal time window size is determined, as described in Section III-B.

Long Short-Term Memory (LSTM) \cite{Hochreiter1997-vm} is used as a deep learning model for robot autonomy to deal with time-series data and is trained based on pre-processed data. The PyTorch library is used to implement the LSTM network. 28 hidden units with 64 batch sizes are used to construct the LSTM. The number of epochs is controlled by the early stopping technique. Six different models are trained for each human subject using six sets of $\Xi_R$ based on the Lipschitz constants between the $100^{th}$ and $50^{th}$ percentile in increments of $10^{th}$ percentiles. Those models are tested on the corresponding human subjects.

\section{Experiment and Results}
The data acquired from three human subjects were processed offline, and the resultant predictors and task mode estimator were implemented on the prototype robot system, as described in the workflow of Fig. \ref{int_vol_re}. Three significant results have been obtained.

\subsection{Reduction of Voluntary Control Efforts with Estimation of Task Mode}
In section \ref{sec:pred_w_task}, we showed that regression predictability has been improved by demonstrating that the Lipschitz quotients are decreased when data is segmented into four modes of the task based on HMM as illustrated in Fig. \ref{hmm_dist}. A t-test is performed to statistically validate the improvement and it is found to be statistically highly significant ($p=$ 5.63e-6 $<$ 0.001) when compared with random segmentation. 

\begin{figure}[t]
\centering
\includegraphics[width=0.42\textwidth]{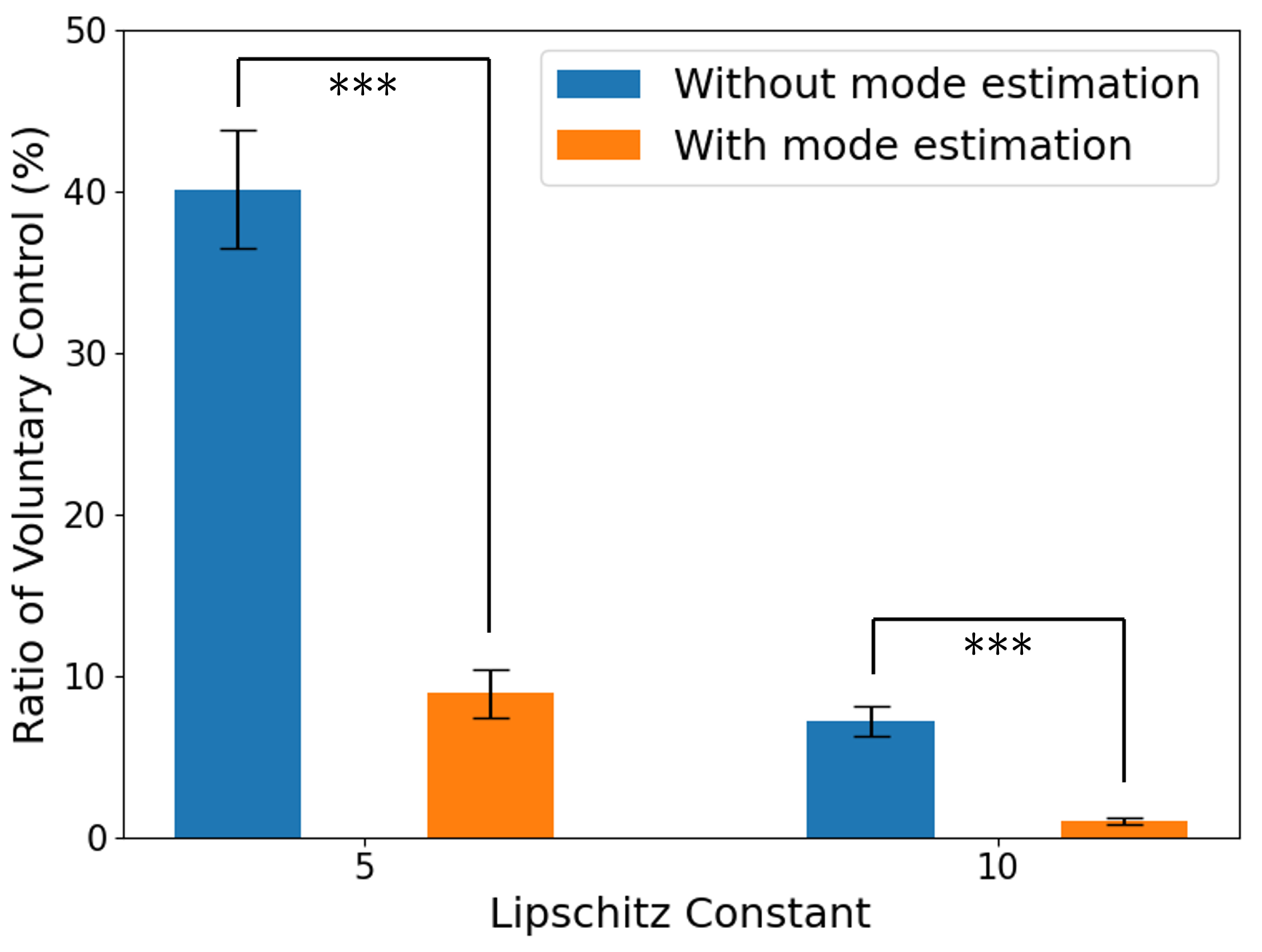}
\caption{The ratio of voluntary control in a single demonstration has decreased from 40.1\% to 8.9\% when task knowledge is incorporated into the predictability analysis with a Lipschitz constant 5. Error bars in the plot represent standard errors. }
\label{fig:task_know_result}
\end{figure}

Smaller Lipschitz quotients should result in less voluntary control; Fewer cases are rendered to voluntary control for the same Lipschitz constant in (\ref{eq:data_mode_vol_re}). Fig. \ref{fig:task_know_result} verified this. The percentage of cases (data points) assigned to voluntary control in a single demonstration data is decreased from 40.1\% to 8.9\% when the Lipschitz constant is set to 5. According to a t-test, the decrement is statistically significant ($p=$1.13e-9 $<$ 0.001). For the Lipschitz constant of 10, the percentage of voluntary control cases decreased from 7.2\% to 1.0\% 
 as the estimation of task mode is used. Note that the percentage of voluntary control depends on the Lipschitz constant. However, voluntary control efforts are reduced for all Lipschitz constants in a reasonable range. The reduction is statistically meaningful, as shown in the figure. 

\subsection{Prediction Error}

Root Mean Square Error (RMSE) is a traditional metric for evaluating prediction performance and predictability. Although it depends on a specific predictor structure and its tuning as well as on the data used for tuning, it provides an overall performance metric. In the current work focusing on data and the Lipschitz-based predictability, the prediction accuracy in terms of RMSE improves as the Lipschitz constant, i.e. the threshold for excluding confounding data, decreases. Fig. \ref{fig:rmse_result} shows that the RMSE of the reactive control decreases as Lipschitz constants are lowered from $100^{th}$ to $50^{th}$ percentile. Here, the percentile means the percentage of data points satisfying the Lipschitz condition with the parameter $K$. Note that RMSE is measured only for the reactive control subset, $\Xi_R$. These results confirm that Lipschitz quotients can also inform prediction accuracy, because excluding data with higher Lipschitz quotients makes prediction easier. 

\subsection{Trade-off between Human Efforts and Autonomy Accuracy}
\begin{figure}[t]
\centering
\includegraphics[width=0.5\textwidth]{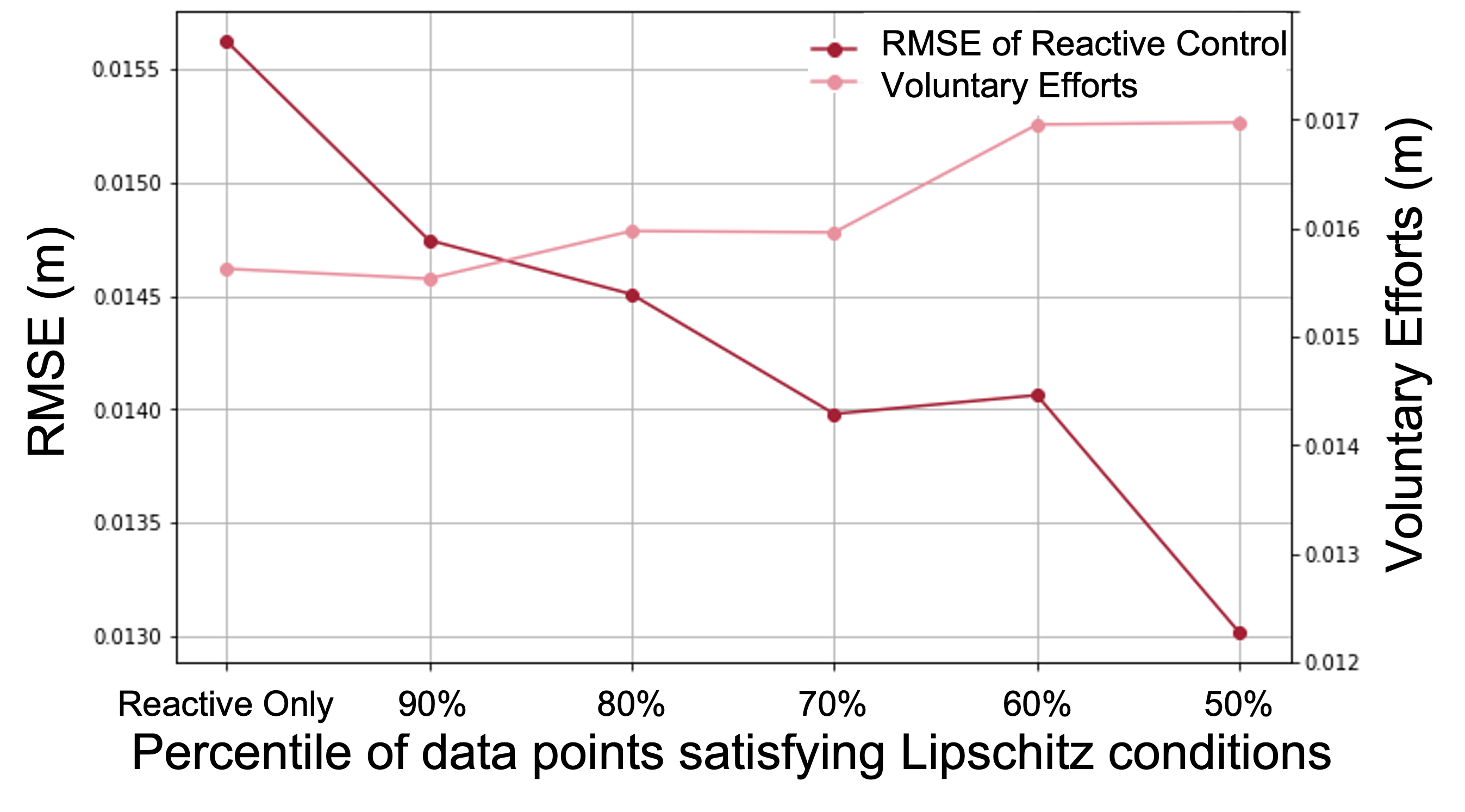}
\caption{RMSE of reactive control decreases and voluntary control efforts increase as lowering the Lipschitz constant.}
\label{fig:rmse_result}
\end{figure}

The other metric to look at is voluntary control efforts, which should be small. The voluntary control efforts are measured by taking the average magnitude of the shoulder motion, i.e. the intended expression of voluntary control.  Fig. \ref{fig:rmse_result} shows that the voluntary control efforts increase as Lipschitz constants are lowered. This is because the lower the Lipschitz constant, the more data is assigned to voluntary control. In contrast, the prediction error decreases as the Lipschitz constant is lowered. There is a trade-off between the performance of the robot autonomy and the amount of human intervention when adjusting their levels based on the Lipschitz constants. The proposed method can provide a quantitative guideline when designing human interaction and robot autonomy in human-robot systems.

\section{Discussion}

\subsection{Harmonized Human-Robot Shared Control}

An effective human-robot shared control system has been constructed with the use of Lipschitz conditions. Confounding situations in the input space are identified through Lipschitz analysis, and additional signals and missing information are sought for differentiating the confounding inputs. This makes the robot's reactive controller more accurate and allows for the performance of the task with higher confidence. Furthermore, human voluntary control is exploited for situations where the Lipschitz quotients are still high even after augmenting the input space. This yields the following features in the human-robot shared control:
\begin{itemize}
    \item The robot reactive control does not interfere or "chip in" with the human voluntary control in a situation where the Lipschitz quotient is high.
    \item The trade-off between prediction accuracy and human control effort can be made by tuning Lipschitz conditions - a threshold of Lipschitz quotients.
\end{itemize}

\subsection{Motions with high Lipschitz quotients}

In section \ref{sec:pred_w_task}, we have seen that the regression predictability is significantly improved after adding a task mode estimate based on the HMM task model. However, we have also found that the Lipschitz quotients remain high for specific situations, such as cutting and collecting. Two reasons can be considered. One is that the added information is still not good enough to differentiate the confounding cases. The other is that human actions are purely spontaneous in a specific situation. In the latter case, there is no regression model that can predict the knife motion in relation to the fork motion. For example, a reciprocating cutting motion of the knife is difficult to generate as a function of the fork motion. During cutting, the fork is holding down a food item and the signals of the fork are almost constant. The motion of the knife, however, is reciprocating, which cannot be generated as a function of the fork's state. The reciprocal motion is spontaneous and is controlled voluntarily by the human. It is rational to leave these cases to human voluntary control.

\subsection{Data Density and Computational Load}
One of the critical assumptions when using Lipschitz conditions is that the data is sufficiently dense. If individual data points are separated in a sparse data space, the denominator of a Lipschitz quotient does not come close to zero. In consequence, the Lipschitz quotient never becomes significantly high, leading to underestimation of the difficulty in prediction and inaccurate predictability assessment.

On the other hand, if the data is dense everywhere, the computation of Lipschitz quotients becomes expensive. In the current work, the number of data is on the order of 1 million. The computational time for computing the Lipschitz quotients is approximately an hour for 100-dimensional input data for each human subject.

\section{Conclusion}

A new data-driven quantitative method is presented to identify and integrate voluntary and reactive control of a robotic system in human-robot collaboration based on regression predictability. Regression predictability is based on the Lipschitz conditions, which inform whether a single continuous function for input-output data exists. If there are confounding situations, where similar input data points indicate different output data points, Lipschitz quotients become large. In an effort to seek additional information that could resolve the confounding situations, task mode information is extracted from the whole data by using HMM. If data with large Lipschitz quotients still exist with added information, they are identified as unpredictable actions and therefore assigned to voluntary control. The rest of the data with small Lipschitz quotients are assigned to reactive control. Finally, the integration of voluntary and reactive control is achieved through the weighted sum of controls based on the estimated predictability.

Bimanual eating with a SuperLimb was used as a human-robot shared control task. Robot autonomy and human intervention were effectively integrated with the use of Lipschitz quotients as a quantitative measure.

\bibliographystyle{IEEEtran}
\bibliography{IEEEabrv,superlimb_hybrid.bib}

\begin{thebibliography}{10}
\providecommand{\url}[1]{#1}
\csname url@samestyle\endcsname
\providecommand{\newblock}{\relax}
\providecommand{\bibinfo}[2]{#2}
\providecommand{\BIBentrySTDinterwordspacing}{\spaceskip=0pt\relax}
\providecommand{\BIBentryALTinterwordstretchfactor}{4}
\providecommand{\BIBentryALTinterwordspacing}{\spaceskip=\fontdimen2\font plus
\BIBentryALTinterwordstretchfactor\fontdimen3\font minus
  \fontdimen4\font\relax}
\providecommand{\BIBforeignlanguage}[2]{{%
\expandafter\ifx\csname l@#1\endcsname\relax
\typeout{** WARNING: IEEEtran.bst: No hyphenation pattern has been}%
\typeout{** loaded for the language `#1'. Using the pattern for}%
\typeout{** the default language instead.}%
\else
\language=\csname l@#1\endcsname
\fi
#2}}
\providecommand{\BIBdecl}{\relax}
\BIBdecl

\bibitem{Sheridan1978-et}
T.~B. Sheridan and W.~L. Verplank, ``\BIBforeignlanguage{en}{Human and computer
  control of undersea teleoperators},'' MIT Man-Machine Systems Laboratory,
  Cambridge, MA, Tech. Rep., Jul. 1978.

\bibitem{Endsley1999-he}
M.~R. Endsley and D.~B. Kaber, ``\BIBforeignlanguage{en}{Level of automation
  effects on performance, situation awareness and workload in a dynamic control
  task},'' \emph{\BIBforeignlanguage{en}{Ergonomics}}, vol.~42, no.~3, pp.
  462--492, Mar. 1999.

\bibitem{Parasuraman2000-vr}
R.~Parasuraman, T.~B. Sheridan, and C.~D. Wickens, ``A model for types and
  levels of human interaction with automation,'' \emph{IEEE Transactions on
  Systems, Man, and Cybernetics - Part A: Systems and Humans}, vol.~30, no.~3,
  pp. 286--297, May 2000.

\bibitem{Beer2014-tu}
J.~M. Beer, A.~D. Fisk, and W.~A. Rogers, ``\BIBforeignlanguage{en}{Toward a
  framework for levels of robot autonomy in human-robot interaction},''
  \emph{\BIBforeignlanguage{en}{J Hum Robot Interact}}, vol.~3, no.~2, pp.
  74--99, Jul. 2014.

\bibitem{Asada1989-ek}
H.~Asada and H.~Izumi, ``Automatic program generation from teaching data for
  the hybrid control of robots,'' \emph{IEEE Trans. Rob. Autom.}, vol.~5,
  no.~2, pp. 166--173, Apr. 1989.

\bibitem{Asada1991-qg}
H.~Asada and S.~Liu, ``Transfer of human skills to neural net robot
  controllers,'' in \emph{Proceedings. 1991 {IEEE} International Conference on
  Robotics and Automation}, Apr. 1991, pp. 2442--2448 vol.3.

\bibitem{Liu1992-pv}
S.~Liu and H.~Asada, ``Transfer of human skills to robots: Learning from human
  demonstrations for building an adaptive control system,'' in \emph{1992
  American Control Conference}.\hskip 1em plus 0.5em minus 0.4em\relax IEEE,
  Jun. 1992, pp. 2607--2612.

\bibitem{He1993-pd}
X.~He and H.~Asada, ``A new method for identifying orders of {Input-Output}
  models for nonlinear dynamic systems,'' in \emph{1993 American Control
  Conference}, Jun. 1993, pp. 2520--2523.

\bibitem{Parietti2017-lm}
F.~Parietti and H.~H. Asada, ``Independent, voluntary control of extra robotic
  limbs,'' in \emph{2017 {IEEE} International Conference on Robotics and
  Automation ({ICRA})}, May 2017, pp. 5954--5961.

\bibitem{Sasaki2017-kq}
T.~Sasaki, M.~Y. Saraiji, C.~L. Fernando, K.~Minamizawa, and M.~Inami,
  ``{MetaLimbs}: multiple arms interaction metamorphism,'' in \emph{{ACM}
  {SIGGRAPH} 2017 Emerging Technologies}, ser. SIGGRAPH '17, no. Article
  16.\hskip 1em plus 0.5em minus 0.4em\relax New York, NY, USA: Association for
  Computing Machinery, Jul. 2017, pp. 1--2.

\bibitem{Vatsal2018-rc}
V.~Vatsal and G.~Hoffman, ``Design and analysis of a wearable robotic
  forearm,'' in \emph{2018 {IEEE} International Conference on Robotics and
  Automation ({ICRA})}, May 2018, pp. 5489--5496.

\bibitem{Veronneau2020-ss}
C.~Veronneau, J.~Denis, L.-P. Lebel, M.~Denninger, V.~Blanchard, A.~Girard, and
  J.-S. Plante, ``\BIBforeignlanguage{en}{Multifunctional remotely actuated
  {3-DOF} supernumerary robotic arm based on magnetorheological clutches and
  hydrostatic transmission lines},'' \emph{\BIBforeignlanguage{en}{IEEE Robot.
  Autom. Lett.}}, vol.~5, no.~2, pp. 2546--2553, Apr. 2020.

\bibitem{Amanhoud2021-rt}
W.~Amanhoud, J.~Hernandez~Sanchez, M.~Bouri, and A.~Billard,
  ``Contact-initiated shared control strategies for four-arm supernumerary
  manipulation with foot interfaces,'' \emph{Int. J. Rob. Res.}, vol.~40, no.
  8-9, pp. 986--1014, Aug. 2021.

\bibitem{Nguyen2019-ky}
P.~H. Nguyen, C.~Sparks, S.~G. Nuthi, N.~M. Vale, and P.~Polygerinos,
  ``\BIBforeignlanguage{en}{Soft {Poly-Limbs}: Toward a new paradigm of mobile
  manipulation for daily living tasks},'' \emph{\BIBforeignlanguage{en}{Soft
  Robot}}, vol.~6, no.~1, pp. 38--53, Feb. 2019.

\bibitem{Bonilla2014-tb}
B.~L. Bonilla and H.~H. Asada, ``A robot on the shoulder: Coordinated
  human-wearable robot control using coloured petri nets and partial least
  squares predictions,'' in \emph{2014 {IEEE} International Conference on
  Robotics and Automation ({ICRA})}, May 2014, pp. 119--125.

\bibitem{Parietti2014-sa}
F.~Parietti and H.~H. Asada, ``Supernumerary robotic limbs for aircraft
  fuselage assembly: Body stabilization and guidance by bracing,'' in
  \emph{2014 {IEEE} International Conference on Robotics and Automation
  ({ICRA})}, May 2014, pp. 1176--1183.

\bibitem{robotics8040102}
\BIBentryALTinterwordspacing
M.~Malvezzi, Z.~Iqbal, M.~C. Valigi, M.~Pozzi, D.~Prattichizzo, and
  G.~Salvietti, ``Design of multiple wearable robotic extra fingers for human
  hand augmentation,'' \emph{Robotics}, vol.~8, no.~4, 2019. [Online].
  Available: \url{https://www.mdpi.com/2218-6581/8/4/102}
\BIBentrySTDinterwordspacing

\bibitem{Cunningham2018-eh}
J.~Cunningham, A.~Hapsari, P.~Guilleminot, A.~Shafti, and A.~A. Faisal,
  ``\BIBforeignlanguage{en}{The supernumerary robotic 3rd thumb for skilled
  music tasks},'' in \emph{\BIBforeignlanguage{en}{2018 7th {IEEE}
  International Conference on Biomedical Robotics and Biomechatronics
  (Biorob)}}.\hskip 1em plus 0.5em minus 0.4em\relax IEEE, Aug. 2018.

\bibitem{Hussain2017-ew}
I.~Hussain, G.~Salvietti, G.~Spagnoletti, M.~Malvezzi, D.~Cioncoloni, S.~Rossi,
  and D.~Prattichizzo, ``A soft supernumerary robotic finger and mobile arm
  support for grasping compensation and hemiparetic upper limb
  rehabilitation,'' \emph{Rob. Auton. Syst.}, vol.~93, pp. 1--12, Jul. 2017.

\bibitem{Prattichizzo2014-tm}
D.~Prattichizzo, M.~Malvezzi, I.~Hussain, and G.~Salvietti, ``The
  {Sixth-Finger}: A modular extra-finger to enhance human hand capabilities,''
  in \emph{The 23rd {IEEE} International Symposium on Robot and Human
  Interactive Communication}, Aug. 2014, pp. 993--998.

\bibitem{Wu2014-qc}
F.~Wu and H.~Asada, ``\BIBforeignlanguage{en}{Supernumerary robotic fingers: An
  alternative {Upper-Limb} prosthesis},'' \emph{\BIBforeignlanguage{en}{ASME
  2014 Dynamic Systems and Control Conference}}, p. V002T16A009, Dec. 2014.

\bibitem{Kieliba2021-vb}
P.~Kieliba, D.~Clode, R.~O. Maimon-Mor, and T.~R. Makin,
  ``\BIBforeignlanguage{en}{Robotic hand augmentation drives changes in neural
  body representation},'' \emph{\BIBforeignlanguage{en}{Sci Robot}}, vol.~6,
  no.~54, May 2021.

\bibitem{Hao2020-sl}
M.~Hao, J.~Zhang, K.~Chen, H.~Asada, and C.~Fu,
  ``\BIBforeignlanguage{en}{Supernumerary robotic limbs to assist human walking
  with load carriage},'' \emph{\BIBforeignlanguage{en}{J. Mech. Robot.}},
  vol.~12, no.~6, p. 061014, Jul. 2020.

\bibitem{Khazoom2020-jw}
C.~Khazoom, P.~Caillouette, A.~Girard, and J.-S. Plante, ``A supernumerary
  robotic leg powered by magnetorheological actuators to assist human
  locomotion,'' \emph{IEEE Robotics and Automation Letters}, vol.~5, no.~4, pp.
  5143--5150, Oct. 2020.

\bibitem{Treers2017-ru}
L.~Treers, R.~Lo, M.~Cheung, A.~Guy, J.~Guggenheim, F.~Parietti, and H.~Asada,
  ``Design and control of lightweight supernumerary robotic limbs for
  {Sitting/Standing} assistance,'' in \emph{2016 International Symposium on
  Experimental Robotics}.\hskip 1em plus 0.5em minus 0.4em\relax Springer
  International Publishing, 2017, pp. 299--308.

\bibitem{Parietti2015-sp}
F.~Parietti, K.~C. Chan, B.~Hunter, and H.~H. Asada, ``Design and control of
  supernumerary robotic limbs for balance augmentation,'' in \emph{2015 {IEEE}
  International Conference on Robotics and Automation ({ICRA})}, May 2015, pp.
  5010--5017.

\bibitem{Kurek2017-sm}
D.~A. Kurek and H.~H. Asada, ``The {MantisBot}: Design and impedance control of
  supernumerary robotic limbs for near-ground work,'' in \emph{2017 {IEEE}
  International Conference on Robotics and Automation ({ICRA})}, May 2017, pp.
  5942--5947.

\bibitem{PLoS-Foot-2015}
E.~Abdi, E.~Burdet, M.~Bouri, and H.~Bleuler, ``\BIBforeignlanguage{en}{Control
  of a supernumerary robotic hand by foot: An experimental study in virtual
  reality},'' \emph{\BIBforeignlanguage{en}{PLoS One}}, vol.~10, no.~7, p.
  e0134501, Jul. 2015.

\bibitem{Huang2020-hc}
Y.~Huang, E.~Burdet, L.~Cao, P.~T. Phan, A.~M.~H. Tiong, and S.~J. Phee, ``A
  {Subject-Specific} {Four-Degree-of-Freedom} foot interface to control a
  surgical robot,'' \emph{IEEE/ASME Trans. Mechatron.}, vol.~25, no.~2, pp.
  951--963, Apr. 2020.

\bibitem{Guggenheim2020-lg}
J.~Guggenheim, R.~Hoffman, H.~Song, and H.~H. Asada, ``Leveraging the human
  operator in the design and control of supernumerary robotic limbs,''
  \emph{IEEE Robotics and Automation Letters}, vol.~5, no.~2, pp. 2177--2184,
  Apr. 2020.

\bibitem{Hussain2016-kn}
I.~Hussain, G.~Spagnoletti, G.~Salvietti, and D.~Prattichizzo,
  ``\BIBforeignlanguage{en}{An {EMG} interface for the control of motion and
  compliance of a supernumerary robotic finger},''
  \emph{\BIBforeignlanguage{en}{Front. Neurorobot.}}, vol.~10, p.~18, Nov.
  2016.

\bibitem{Wu2016-xf}
F.~Y. Wu and H.~Harry~Asada, ``Implicit and intuitive grasp posture control for
  wearable robotic fingers: A {Data-Driven} method using partial least
  squares,'' \emph{IEEE Trans. Rob.}, vol.~32, no.~1, pp. 176--186, Jan. 2016.

\bibitem{Setiawan2020-yk}
J.~D. Setiawan, M.~Ariyanto, M.~Munadi, M.~Mutoha, A.~Glowacz, and
  W.~Caesarendra, ``\BIBforeignlanguage{en}{Grasp posture control of wearable
  extra robotic fingers with flex sensors based on neural network},''
  \emph{\BIBforeignlanguage{en}{Electronics (Basel)}}, vol.~9, no.~6, p. 905,
  May 2020.

\bibitem{Song2021-dh}
H.~Song and H.~H. Asada, ``Integrated {Voluntary-Reactive} control of a
  {Human-SuperLimb} hybrid system for hemiplegic patient support,'' \emph{IEEE
  Robotics and Automation Letters}, vol.~6, no.~2, pp. 1646--1653, Apr. 2021.

\bibitem{Pomerleau-1989-15721}
D.~Pomerleau, ``Alvinn: An autonomous land vehicle in a neural network,'' in
  \emph{Proceedings of (NeurIPS) Neural Information Processing Systems},
  D.~Touretzky, Ed.\hskip 1em plus 0.5em minus 0.4em\relax Morgan Kaufmann,
  December 1989, pp. 305 -- 313.

\bibitem{Ross2011-gj}
S.~Ross, G.~Gordon, and D.~Bagnell, ``A reduction of imitation learning and
  structured prediction to {No-Regret} online learning,'' in \emph{Proceedings
  of the Fourteenth International Conference on Artificial Intelligence and
  Statistics}, ser. Proceedings of Machine Learning Research, G.~Gordon,
  D.~Dunson, and M.~Dud{\'\i}k, Eds., vol.~15.\hskip 1em plus 0.5em minus
  0.4em\relax Fort Lauderdale, FL, USA: PMLR, 2011, pp. 627--635.

\bibitem{Zhang2018-dy}
T.~Zhang, Z.~McCarthy, O.~Jow, D.~Lee, X.~Chen, K.~Goldberg, and P.~Abbeel,
  ``Deep imitation learning for complex manipulation tasks from virtual reality
  teleoperation,'' in \emph{2018 {IEEE} International Conference on Robotics
  and Automation ({ICRA})}, May 2018, pp. 5628--5635.

\bibitem{Florence2020-mg}
P.~Florence, L.~Manuelli, and R.~Tedrake, ``{Self-Supervised} correspondence in
  visuomotor policy learning,'' \emph{IEEE Robotics and Automation Letters},
  vol.~5, no.~2, pp. 492--499, Apr. 2020.

\bibitem{florence2021implicit}
P.~Florence, C.~Lynch, A.~Zeng, O.~Ramirez, A.~Wahid, L.~Downs, A.~Wong,
  J.~Lee, I.~Mordatch, and J.~Tompson, ``Implicit behavioral cloning,''
  \emph{Conference on Robot Learning (CoRL)}, 2021.

\bibitem{9104757}
A.~Zeng, S.~Song, J.~Lee, A.~Rodriguez, and T.~Funkhouser, ``Tossingbot:
  Learning to throw arbitrary objects with residual physics,'' \emph{IEEE
  Transactions on Robotics}, vol.~36, no.~4, pp. 1307--1319, 2020.

\bibitem{Jacobsson1996-pb}
C.~Jacobsson, K.~Axelsson, B.~I. Wenngren, and A.~Norberg,
  ``\BIBforeignlanguage{en}{Eating despite severe difficulties: assessment of
  poststroke eating},'' \emph{\BIBforeignlanguage{en}{J. Clin. Nurs.}}, vol.~5,
  no.~1, pp. 23--31, Jan. 1996.

\bibitem{Perry2003-rc}
L.~Perry and S.~McLaren, ``\BIBforeignlanguage{en}{Eating difficulties after
  stroke},'' \emph{\BIBforeignlanguage{en}{J. Adv. Nurs.}}, vol.~43, no.~4, pp.
  360--369, Aug. 2003.

\bibitem{Jacobsson2000-so}
C.~Jacobsson, K.~Axelsson, P.~O. Osterlind, and A.~Norberg,
  ``\BIBforeignlanguage{en}{How people with stroke and healthy older people
  experience the eating process},'' \emph{\BIBforeignlanguage{en}{J. Clin.
  Nurs.}}, vol.~9, no.~2, pp. 255--264, Mar. 2000.

\bibitem{Rabiner1989-me}
L.~R. Rabiner, ``A tutorial on hidden markov models and selected applications
  in speech recognition,'' \emph{Proc. IEEE}, vol.~77, no.~2, pp. 257--286,
  Feb. 1989.

\bibitem{Bilmes1998-bq}
J.~Bilmes, ``\BIBforeignlanguage{en}{A gentle tutorial of the em algorithm and
  its application to parameter estimation for gaussian mixture and hidden
  markov models},'' \emph{\BIBforeignlanguage{en}{CTIT technical reports
  series}}, 1998.

\bibitem{Baum1972AnIA}
L.~E. Baum, ``An inequality and associated maximization technique in
  statistical estimation of probabilistic functions of a markov process,''
  1972.

\bibitem{Dempster1977-ze}
A.~P. Dempster, N.~M. Laird, and D.~B. Rubin, ``\BIBforeignlanguage{en}{Maximum
  likelihood from incomplete data via {theEMAlgorithm}},''
  \emph{\BIBforeignlanguage{en}{J. R. Stat. Soc.}}, vol.~39, no.~1, pp. 1--22,
  Sep. 1977.

\bibitem{Viterbi1967-vu}
A.~Viterbi, ``\BIBforeignlanguage{en}{Error bounds for convolutional codes and
  an asymptotically optimum decoding algorithm},''
  \emph{\BIBforeignlanguage{en}{IEEE Trans. Inf. Theory}}, vol.~13, no.~2, pp.
  260--269, Apr. 1967.

\bibitem{Reeves1975-st}
B.~Reeves, ``\BIBforeignlanguage{en}{The natural history of the frozen shoulder
  syndrome},'' \emph{\BIBforeignlanguage{en}{Scand. J. Rheumatol.}}, vol.~4,
  no.~4, pp. 193--196, 1975.

\bibitem{Lindgren2007-oq}
I.~Lindgren, A.-C. J{\"o}nsson, B.~Norrving, and A.~Lindgren, ``Shoulder pain
  after stroke,'' \emph{Stroke}, vol.~38, no.~2, pp. 343--348, Feb. 2007.

\bibitem{Hochreiter1997-vm}
S.~Hochreiter and J.~Schmidhuber, ``Long {Short-Term} memory,'' \emph{Neural
  Comput.}, vol.~9, no.~8, pp. 1735--1780, Nov. 1997.

\end{thebibliography}
\end{document}